\documentclass{article}

\usepackage{arxiv}

\usepackage[utf8]{inputenc} % allow utf-8 input

\usepackage{amsfonts}       % blackboard math symbols
\usepackage{nicefrac}       % compact symbols for 1/2, etc.
\usepackage{microtype}      % microtypography
\usepackage{lipsum}

\usepackage{color}
\usepackage{listings}
\usepackage{booktabs}
\usepackage{array}
\usepackage{graphicx}
\usepackage{longtable}
\usepackage{subfigure}

\usepackage{cite}
\usepackage{url}
\usepackage{fancyhdr}

\usepackage{soul}

\usepackage{float}
\usepackage{listings}
\usepackage{mdwmath}
\usepackage{mdwtab}
\usepackage{multirow}
\usepackage{multicol}
\usepackage{rotating}
\usepackage{setspace}
\usepackage[utf8]{inputenc}
\usepackage{lineno}
\usepackage{listings}

\usepackage{mdwmath}
\usepackage{mdwtab}
\usepackage{multirow}
\usepackage{multicol}
\usepackage{array}
\usepackage{booktabs}

\usepackage{enumitem}
\usepackage{xspace}
\usepackage[export]{adjustbox}
\usepackage{graphicx}
\usepackage{color,soul}
\usepackage{rotating}
\usepackage{setspace}
\usepackage{amsmath} 
\usepackage{amssymb}
\usepackage{float}
\usepackage[dvipsnames]{xcolor}

\usepackage{hyperref}
\usepackage{url}

\title{Agent Design Pattern Catalogue: \\A Collection of Architectural Patterns \\ for Foundation Model based Agents}
\author{Yue Liu, Sin Kit Lo, Qinghua Lu, Liming Zhu, Dehai Zhao, Xiwei Xu, Stefan Harrer, Jon Whittle \\
Data61, CSIRO, Australia \\
Email: \textit{firstname.lastname}@data61.csiro.au
}
\date{May 2024}

\begin{document}

\maketitle

Foundation model-enabled generative artificial intelligence facilitates the development and implementation of agents, which can leverage distinguished reasoning and language processing capabilities to takes a proactive, autonomous role to pursue users' goals. Nevertheless, there is a lack of systematic knowledge to guide practitioners in designing the agents considering challenges of goal-seeking (including generating instrumental goals and plans), such as hallucinations inherent in foundation models, explainability of reasoning process, complex accountability, etc. To address this issue, we have performed a systematic literature review to understand the state-of-the-art foundation model-based agents and the broader ecosystem. In this paper, we present a pattern catalogue consisting of 18 architectural patterns with analyses of the context, forces, and trade-offs as the outcomes from the previous literature review. We propose a decision model for selecting the patterns. The proposed catalogue can provide holistic guidance for the effective use of patterns, and support the architecture design of foundation model-based agents by facilitating goal-seeking and plan generation.

\section{Introduction}
Being the technical backbones of the highly disruptive generative artificial intelligence (GenAI) technologies, foundation models (FMs) have received a vast amount of attention from academia and industries~\cite{bommasani2021opportunities}. Specifically, the emergence of large language models (LLMs) with their remarkable capabilities to understand and generate human-like reasoning and content has sparked the growth of a diverse range of downstream tasks using language models. Subsequently, there is a rapidly growing interest in the development of FM-based autonomous agents, e.g., AutoGPT\footnote{\url{https://github.com/Significant-Gravitas/AutoGPT}\label{AutoGPT}} and BabyAGI\footnote{\url{https://github.com/yoheinakajima/babyagi}}, which can take a proactive, autonomous role to pursue users' goals. This goal could be broad given by human, necessitating the agents to derive their autonomy from the capabilities of FMs, enabling them to segregate the goal into a set of executable tasks and orchestrate task execution to fulfill the goal. During the reasoning process, humans can also provide feedback on instrumental goals, revise a multi-step plan derived by the agent, correct intermediate results, or even refine a plan/goal during execution based on early outcomes. 

%However, FMs exhibit limitations, particularly in understanding and performing complex chains of tasks. This triggers 

%With autonomous agents, users provide only a high-level goal, without explicit step-by-step instructions, while these agents derive their autonomy from the capabilities of FMs, enabling them to segregate the goal into a set of executable tasks and orchestrate task execution to fulfill the goal. 

While huge efforts have been put into this merging field, there is a steep learning curve for practitioners to build and implement FM-based agents. We noticed that there are a series of reusable solutions that can be grouped into patterns to address the diverse challenges in designing FM-based agents, however, the architecture design and architectural patterns collection of the agents have not been systematically explored and formulated. Furthermore, the design of systems that integrate agents is non-trivial and complex, especially in how to select appropriate design decisions to fulfill different software quality requirements and design constraints. Further, multi-agent systems may require additional considerations on the coordination and interactions of agents, for instance, collusion between agents, and correlated failures~\cite{anwar2024foundational}. We list several challenges in developing and implementing FM-based agents as follows:

\begin{itemize}

    \item Agents often struggle to fully comprehend and execute complex tasks, leading to the potential for inaccurate responses. This challenge may be intensified by the inherent reasoning uncertainties during plan generation and action procedures. For instance, across a long-term planning, the included steps may depend on each other, even slight deviation to a few steps can significantly impact the overall success rate.

    \item Agents should not be entirely blamed for inaccurate response, since users may provide limited context, ambiguous goals or unclear instructions during the interaction with agents, which will result in underspecification~\cite{10.1145/3593013.3594033, JMLR:v23:20-1335} in the reasoning process and response generation of agents.

    \item The sophisticated internal architecture of agents and foundation models results in limited explainability, making them ``black boxes'' to stakeholders. Consequently, agents often struggle to interpret their reasoning steps, which can affect the reliability, robustness, and overall trustworthiness of agent systems.
    
    %Lack of analysis of the overall ecosystem consisting of different stakeholders, FM-based agents, non-agent AI models, and non-AI software applications.
    
    \item The accountability process is complicated due to the interactions between various stakeholders, FM-based agents, non-agent AI models, and non-AI software applications within the overall ecosystem. Highly autonomous agents may delegate or even create other agents or tools for certain tasks. In this circumstance, responsibility and accountability may be intertwined among multiple entities.

    %responsibilities of different stakeholders, FM-based agents, non-agent AI models, and non-AI software applications, and the interactions between them.
    
    %Lack of proper curation of the challenges, forces, and trade-offs for the reusable solutions in agent development.
    %Many reusable solutions for different challenges in agents implementation but not curated properly.
\end{itemize}

In this regard, we present a catalogue of patterns for foundation model-based agents in this paper, aiming to address the identified issues via providing a holistic guidance to the design and development of different types of agents, and specifying different collaboration schemes between these agents. For instance, the goal creator patterns can clarify users' intentions and instructions to avoid underspecification. A series of patterns for reflection can help identify and mitigate the uncertainties in agent-generated plans, while the explainability of agent reasoning process is improved by requesting an agent to reflect on its generated plan. Accountability can be preserved when agents participate in a vote where their identities and operations are all logged. Please note that ``agent'' can be referred to i) AI acting on behalf of another entity, or; ii) AI that can take active roles or produces effect to achieve users' goals. The former circumstance requires thorough analysis on governance perspective, while hereby, we claim that in this study, we focus on the second concept of ``agents'' that are capable of goal-seeking and plan generation. In software engineering, an architectural pattern is a reusable solution to a problem that occurs commonly within a given context in software design. Our pattern catalogue includes 18 patterns that were identified based on the study conducted by Lu et al.~\cite{lu2024responsible}. The intended audience of collected patterns is software architects and developers who are interested in FM-based agent design and implementation. The contributions of this paper include:

%a person based on strictly predefined rules with no goal-seeking or planning involved, or; ii) AI that can devise the sub-goals from a high-level goal provided by users. 

\begin{itemize}
    \item The collection of architectural patterns provides a design solution pool for practitioners to select from for real-world agent implementations. For instance, architects can apply \textit{passive goal creator} or \textit{proactive goal creator} considering the application scenarios and the requirements for accessibility.

    \item The FM-based agent ecosystem with architectural pattern annotations, serving as a guide for the design and development of FM-based agents. In particular, an agent can request feedback from both human and other agents for reasoning certainty and improved explainability, and there are three cooperation schemes for multiple-agent systems with different accountability processes.

    \item The curated analysis of each included pattern regarding the application context, addressed issues, consequent benefits and trade-offs on software quality attributes, real-world known uses, and the relationship with other patterns.

    \item A decision model that can help architects structure the included patterns and make rational design decisions on foundation model based agents. We also share the experiences on pattern application in different research projects.
    
\end{itemize}

The remainder of this paper is organised as follows. Section~\ref{sec:background} introduces background knowledge and discusses related work. The methodology of this research is introduced in Section~\ref{sec:methodology} and Section~\ref{sec:pattern} presents each pattern in detail with our extended pattern template. Section~\ref{sec:discussion} illustrates a decision model for pattern selection, and discusses the insights we obtained in this research project, while Section~\ref{sec:conclusion} concludes the paper.

\section{Background \& Related Work}
\label{sec:background}

The introduction of OpenAI’s ChatGPT~\cite{openai2024gpt4} in November 2022 has gathered over 100 million users in two months upon its release, becoming the fastest-growing consumer internet app of all time~\cite{Hu_2023}. This has also initiated the race among big tech arms in the development of FM and GenAI products. For instance, Google rolled out its own GenAI product, the Bard models~\footnote{\url{https://ai.google/static/documents/google-about-bard.pdf}}, then released the Gemini~\footnote{\url{https://gemini.google.com/}\label{gemini}} as the updated version. Anthropic has also emerged as one of the major FM providers since the launch of their Claude models~\footnote{\url{https://www.anthropic.com/claude}} There are also many open-sourced FMs, such as Llama~\footnote{\url{https://llama.meta.com/}} and Mistral~\footnote{\url{https://mistral.ai/}}. Schulhoff et al.~\cite{schulhoff2024prompt} performed a systematic literature review and proposed a taxonomy of diverse prompting techniques for foundation models.

With the explosive growth of FMs, it is highly notable that FM-based agents come into the picture. AI agents are typically designed to operate a particular software environment. A single agent is able to take actions in a variety of three-dimensional virtual worlds~\cite{DeepLearningAI_2024}. Recently, there have been a lot of studies that present the architecture of their agents and broader AI systems~\cite{packer2024memgpt, COLABIANCHI2023100510}. LangChain analysed the cognitive architecture of agents~\cite{Cognitive_Architecture}. However, these architectures often focus only on certain components or schemes. For instance, Packer et al.~\cite{packer2024memgpt} explicitly covered the memory design of the agent. Andrew Ng discussed reflection, tool use, planning and multi-agent collaboration~\cite{DeepLearningAI_2024} and provided a corresponding course\footnote{\url{https://www.deeplearning.ai/short-courses/ai-agentic-design-patterns-with-autogen/}}. Jain presented four design patterns for compound AI systems, including retrieval augmented generation, conversational AI, multi-agent communication, and co-pilot~\cite{Jain_pattern}. Zhou et al.~\cite{10.1145/3686802} proposed a four-tiered hierarchical artificial society model for the ``cyber-physical-social'' aspects of adaptive AI agents: i) an autonomous layer for agent memory, behavior and decision, ii) an evolutionary layer for agent learning and heterogeneity, iii) an interactive layer for agent collaboration, competition and topological structure, and iv) an emergent layer for the overall environment, feedback, and intervention. Yan et al.~\cite{Year_of_Building_with_LLMs} shared their experience of building products with LLMs and the best practices they summarised during this process. Hassan et al.~\cite{hassan2024rethinking} demonstrated a high-level structure of FMware, consisting of Agentware and Promptware, and identified the challenges of software engineering for FMware. Gao et al.~\cite{gao2024empowering} proposed the roadmap for designing biomedical AI agents, and illustrated the required components.

%https://docs.anthropic.com/claude/docs/tool-use
%tool use

We noticed that there is a lack of a holistic view of architecture design, making it challenging to develop FM-based agents. Moreover, software architecture comprises software elements, relations among them, and properties of both~\cite{10.5555/2392670}. With the increase in the incorporation of machine learning into software and systems, methods to identify the impact on the reliability of machine learning are essential to ensure the reliability of the software and systems in which these algorithms reside~\cite{9153718}. Hence, frameworks that only list the high-level components supporting their functionality usually lack system-level thinking, with no explicit identification of software components, relationships among them, and their properties~\cite{lu2024responsible}. Our work presents 18 design patterns regarding agent goal-seeking and plan generation. The pattern catalogue can provide a holistic guidance to practitioners on the trade-off analysis for the design of foundation model-based agents.

\begin{figure*}[t]
    \centering
    \includegraphics[width=\columnwidth]{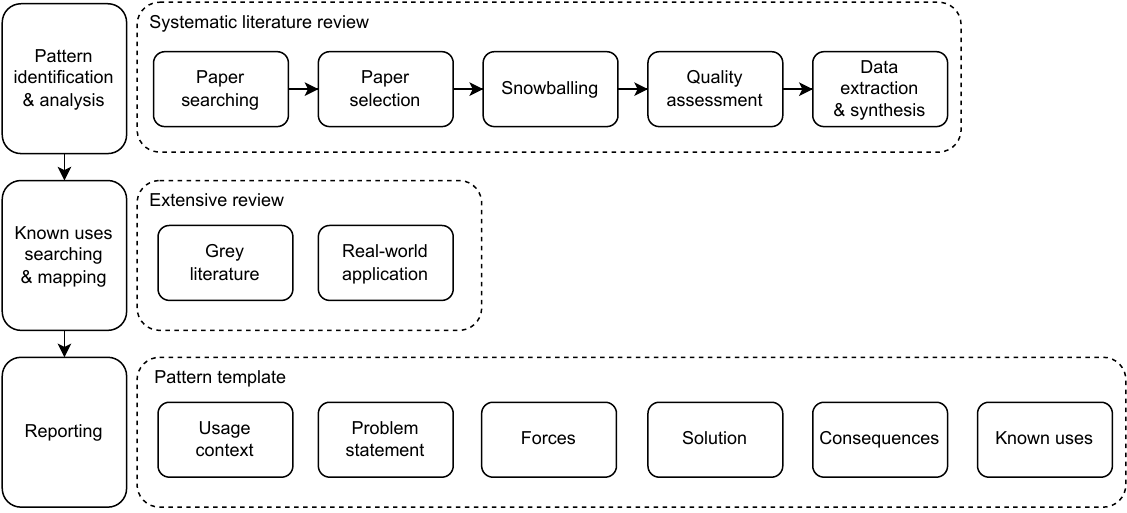}	
    \caption{Methodology.}
    \label{fig:methodology}
\end{figure*}

\begin{figure*}[t]
    \centering
    \includegraphics[width=\columnwidth]{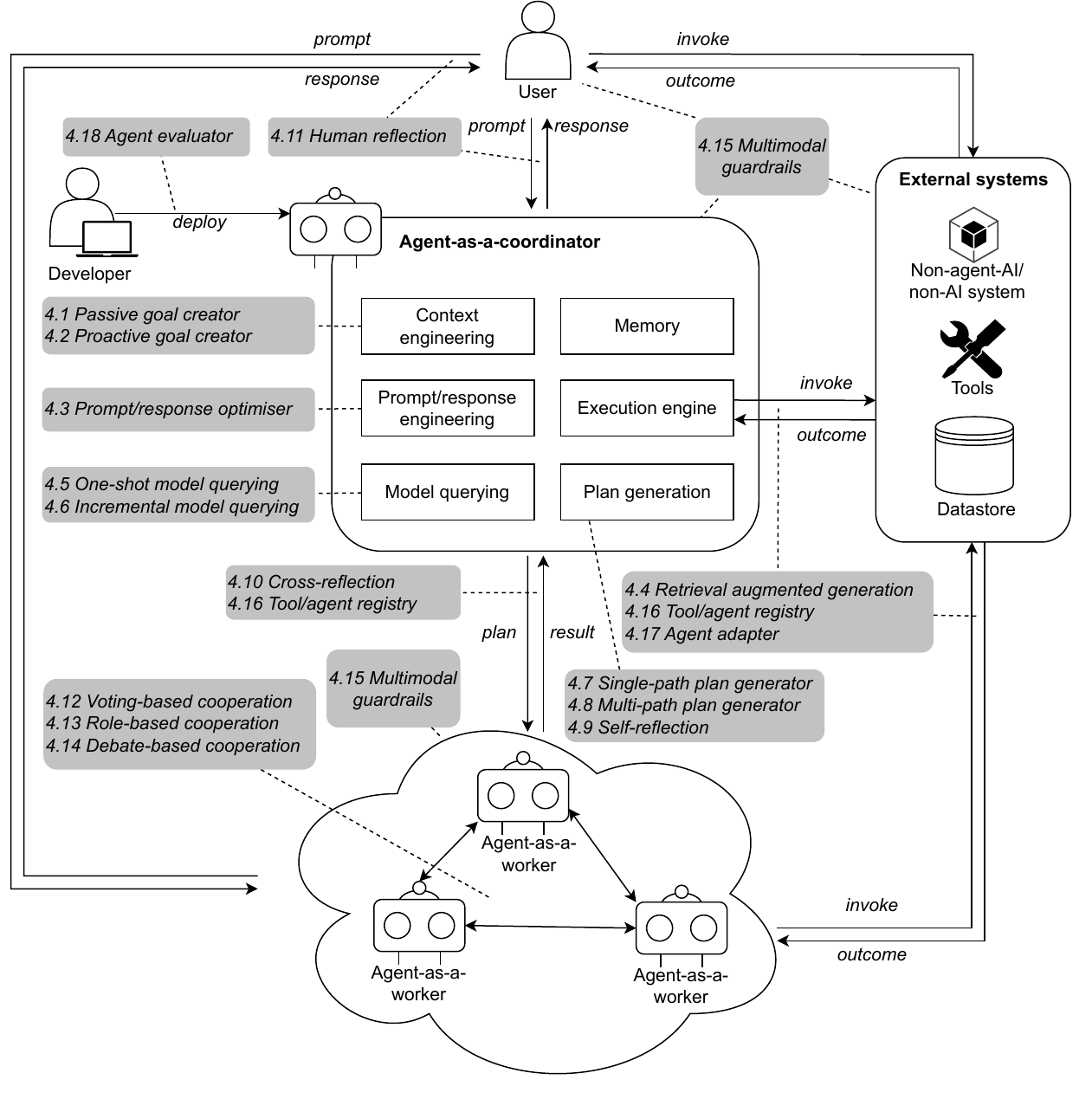}	
    \caption{Ecosystem of FM-based agent systems annotated with architectural patterns in gray boxes.}
    \label{fig:ecosystem}
\end{figure*}

\section{Methodology}
\label{sec:methodology}

Fig.~\ref{fig:methodology} illustrates the pattern extraction and collection process. First, we conducted a systematic literature review (SLR) on FM-based agents~\cite{lu2024responsible}. We focused on available materials and research works that are highly academic-based. We selected relevant papers based on a series of preset criteria and conducted both forward and backward snowballing processes to identify materials that were left out. After finalising the paper pool, we performed quality assessments on the selected materials to ensure the quality of the work. Finally, 57 studies were included for data extraction and synthesis. A pattern-oriented reference architecture for foundation model-based agents was proposed based on the findings.

Based on the reported findings, we delved into the analysis of identified patterns for building integrating FM-based agents. Through the SLR, we have identified a series of architectural design challenges in the development and implementation of systems with the integration of agents. We then further conducted extensive review on this topic, which includes grey literature, and real-world applications (through scrutinising official websites and documents available), for identifying the known uses as the implementation of our included patterns. Combining the findings of both SLR and the additional review, in this paper, we report our findings on 18 extracted patterns.

\begin{table*}[tbp]
\footnotesize
\centering
\caption{Foundation model based agent design pattern catalogue overview.}
\label{tab:overview}
%\begin{spacing}{1} 
\begin{tabular}{p{0.18\columnwidth}p{0.75\columnwidth}}
\toprule

\multicolumn{1}{c}{\bf Pattern} &
\multicolumn{1}{c}{\bf Summary}\\
\midrule

\multirow{2}{0.18\columnwidth}{Passive goal creator} & Analyse users' articulated prompts through the dialogue interface to preserve \textit{interactivity}, \textit{goal-seeking} and \textit{efficiency}.\\
\cmidrule(l){1-2}

\multirow{2}{0.18\columnwidth}{Proactive goal creator} & Anticipate users' goals by understanding human interactions and capturing the context via relevant tools, to enhance \textit{interactivity}, \textit{goal-seeking} and \textit{accessibility}. \\
\cmidrule(l){1-2}

\multirow{2}{0.18\columnwidth}{Prompt/response optimiser} & Optimise the prompts/responses according to the desired input or output content and format to provide \textit{standardisation}, \textit{goal alignment}, \textit{interoperability} and \textit{adaptability}. \\
\cmidrule(l){1-2}

\multirow{2}{0.18\columnwidth}{Retrieval augmented generation} & Enhance the knowledge \textit{updatability} of the agents while maintaining \textit{data privacy} of on-premise foundation model-based agents/systems implementations. \\
\cmidrule(l){1-2}

\multirow{2}{0.18\columnwidth}{One-shot model querying} & Access the foundation model in a single instance to generate all necessary steps for the plan for \textit{cost efficiency} and \textit{simplicity}. \\
\cmidrule(l){1-2}

\multirow{2}{0.18\columnwidth}{Incremental model querying} & Access the foundation model at each step of the plan generation process to provide \textit{supplementary context}, improve \textit{reasoning certainty} and \textit{explainability}.\\
\cmidrule(l){1-2}

\multirow{2}{0.18\columnwidth}{Single-path plan generator} & Orchestrate the generation of intermediate steps leading to the achievement of the user’s goal to improve \textit{reasoning certainty}, \textit{coherence} and \textit{efficiency}. \\
\cmidrule(l){1-2}

\multirow{2}{0.18\columnwidth}{Multi-path plan generator} & Allow multiple choice creation at each intermediate step leading to achieving users' goals to enhance \textit{reasoning certainty}, \textit{coherence}, \textit{alignment to human preference} and \textit{inclusiveness}. \\
\cmidrule(l){1-2}

\multirow{3}{0.18\columnwidth}{Self-reflection} & Enable the agent to generate feedback on the plan and reasoning process and provide refinement guidance from themselves to improve \textit{reasoning certainty}, \textit{explainability}, \textit{continuous improvement} and \textit{efficiency}. \\
\cmidrule(l){1-2}

\multirow{2}{0.18\columnwidth}{Cross-reflection} & Use different agents or foundation models to provide feedback and refine the generated plan and reasoning process for better \textit{reasoning certainty}, \textit{explainability}, \textit{inclusiveness} and \textit{scalability}. \\
\cmidrule(l){1-2}

\multirow{2}{0.18\columnwidth}{Human reflection} & Collect feedback from humans to refine the plan and reasoning process, to effectively align with \textit{human preference}, improving \textit{contestability} and \textit{effectiveness}. \\
\cmidrule(l){1-2}

\multirow{2}{0.18\columnwidth}{Voting-based cooperation} & \multirow{2}{0.75\columnwidth}{Enable free opinions expression across agents and reach consensus by submitting their votes to preserve \textit{fairness}, \textit{accountability} and \textit{collective intelligence}.}\\ \\
\cmidrule(l){1-2}

\multirow{2}{0.18\columnwidth}{Role-based cooperation} & Assign assorted roles, and finalise decisions in accordance with the roles of agents for to facilitate \textit{division of labor}, \textit{fault tolerance}, \textit{scalability} and \textit{accountability}. \\
\cmidrule(l){1-2}

\multirow{3}{0.18\columnwidth}{Debate-based cooperation} & Provide and receive feedback across multiple agents adjusts the thoughts and behaviors during the debate with other agents until a consensus is reached to improve \textit{adaptability}, \textit{explainability} and \textit{critical thinking}. \\
\cmidrule(l){1-2}

\multirow{3}{0.18\columnwidth}{Multimodal guardrails} & Control the inputs and outputs of foundation models to meet specific requirements such as user requirements, ethical standards, and laws to enhance \textit{robustness}, \textit{safety}, \textit{standard alignment}, and \textit{adaptability}. \\
\cmidrule(l){1-2}

\multirow{2}{0.18\columnwidth}{Tool/agent registry} & Maintain a unified and convenient source to select diverse agents and tools to improve \textit{discoverability}, \textit{efficiency}, \textit{tool appropriateness} and \textit{scalability}. \\
 \cmidrule(l){1-2}

% \multirow{2}{0.18\columnwidth}{Evolutionary Model Merging} & Integrate multiple foundation models into one single model with combined capabilities, enhancing \textit{adaptability}, \textit{fairness}, \textit{response accuracy}, \textit{cost-efficiency} and \textit{scalability}. \\

\multirow{2}{0.18\columnwidth}{Agent adapter} & Provide interface to connect the agent and external tools for task completion, ensuring \textit{interoperability} and \textit{adaptability}, and \textit{reduce development cost}. \\
 \cmidrule(l){1-2}

\multirow{2}{0.18\columnwidth}{Agent evaluator} & Perform testing to assess the agent regarding diverse requirements and metrics, ensuring the \textit{functional suitability}, \textit{adaptability} with improved \textit{flexibility}. \\

\bottomrule
\end{tabular}
%\end{spacing}
\end{table*}

\section{Pattern Catalogue for Foundation Model-based Agents}
\label{sec:pattern}

In this section, we present a pattern catalogue for FM-based agents by adopting the extended pattern template in \cite{patternLanguage}. It includes the pattern name, a short summary, usage context, a problem statement, a discussion on the forces leading to the problem difficulty, the solution and its consequences, and several examples of real-world known uses of the pattern. Please note that for each included pattern, we provide a simplified graphical representation that highlights only the essential components necessary to explain pattern application. Detailed interactions between all agent components have been omitted for clarity. Table~\ref{tab:overview} offers an overview of the collected patterns. 

Fig.~\ref{fig:ecosystem} illustrates the ecosystem of foundation model-based agents, the agent components and interactions between different entities are annotated with the relevant patterns. When users interact with the agent, \textit{passive goal creator} and \textit{proactive goal creator} can help comprehend users' intentions and environmental information, and formalised the eventual goals in context engineering, while \textit{prompt/response optimiser} refines the prompts or instructions to other agents/tools based on the predefined templates for certain format or content requirements. Given users' input, the agent fetches additional context information from the knowledge base via \textit{retrieval augmented generation}. Then, it constructs plans to decompose the ultimate goals into actionable tasks through \textit{single-path plan generator} and \textit{multi-path plan generator}. In this process, \textit{one-shot model querying} and \textit{incremental model querying} may be carried out.

A generated plan should be reviewed to ensure its accuracy, usability, completeness, etc. \textit{Self-refection, cross-reflection,} and \textit{human reflection} can help the agent to collect feedback from different reflective entities, and refine the plan and reasoning steps accordingly. Afterwards, the agent can assign tasks to other narrow AI-based or non-AI systems, invoke external tools, and employ a set of agents for goal achievement by \textit{tool/agent registry}. In particular, agents can work on the same task and finalise the results with \textit{voting-based, role-based,} or \textit{debate-based cooperation}. For instance, agents can act as different roles such as coordinator and worker. \textit{Agent adapter} keeps learning the interfaces of different tools, and convert them into FM-friendly environment. \textit{Multimodal guardrails} can be applied to manage and control the inputs/outputs of foundation models. Meanwhile, the employed agents will conduct respective reasoning, planning and execution process, which may require external systems via \textit{retrieval augmented generation, tool/agent registry,} and \textit{agent adapter} either. Please note that we omit the detailed architecture of agent-as-a-worker, and pattern application in several interactions for the clarity of this diagram, for instance, each agent-as-a-worker has its \textit{passive/proactive goal creator}, \textit{prompt/response optimiser}, and \textit{single/multi-path plan generator}, etc. Finally, we claim that developers can evaluate the performance of agents at both design-time and runtime via \textit{agent evaluator}.

%Finally, the underlying foundation model of each agent can be updated via \textit{evolutionary model merging}, while \textit{multimodal guardrails} can be applied to manage and control the inputs/outputs of foundation models. Please note that we omit the detailed architecture of agent-as-a-worker, and pattern application in several interactions for the simplicity of this diagram, for instance, the invocation of non-agent-AI/non-AI systems and external tools services can all apply \textit{tool/agent registry}, and agent-as-a-worker can all have their respective datastore or knowledge base.

% figure + table + pattern template

\subsection{Passive Goal Creator}

\vspace{0.5em}\noindent \textbf{Summary:} Passive goal creator analyses users' articulated goals through the dialogue interface.

\vspace{0.5em}\noindent \textbf{Context:} When querying agents to address certain issues, users provide related context and explain the goals in prompts.

%Users explain the goals that the agent is expected to achieve in the prompt.

\vspace{0.5em}\noindent \textbf{Problem:} Users may lack expertise of interacting with agents, and the provided information can be ambiguous for goal achievement.

%The provided information may be narrow. How can an agent 

%The agent may generate inaccurate responses to users' instructions when given insufficient context.

\vspace{0.5em}\noindent \textbf{Forces:} 

\begin{itemize}
  %\item \textit{Interactivity.} Users need to interact with the agent to provide instructions and receive responses.

  %\item \textit{Goal-seeking.} The agent requires as much information as possible to understand users' goals.

  %\item \textit{Intuitiveness.} Simple tasks should be assigned and reported in an intuitive manner.

  \item \textit{Underspecification.} Users may not be able to provide thorough context information and specify precise goals to agents.

  \item \textit{Efficiency.} Users expect quick responses from agents.
\end{itemize}

\begin{figure}[!ht]
    \centering
    \includegraphics[width=0.5\columnwidth]{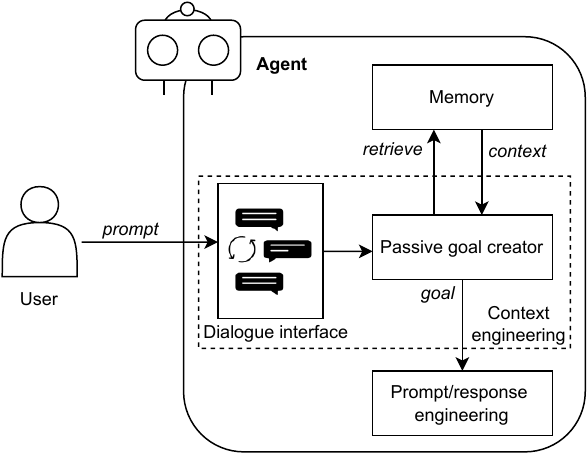}	
    \caption{Passive goal creator.}
    \label{fig:passive}
\end{figure}

\vspace{0.5em}\noindent \textbf{Solution:} Fig.~\ref{fig:passive} illustrates a simple graphical representation of \textit{passive goal creator}. A foundation model-based agent provides a dialogue interface where users can directly specify the context and problems, which are transferred to \textit{passive goal creator} for goal determination. Meanwhile, the \textit{passive goal creator} can also retrieve related information from memory, including the repository of artefacts being worked on, relevant tools used in recent tasks, conversation histories, and the positive and negative examples, which are appended to the user's prompt for goal-seeking. The generated goals are sent to other components for further task decomposition and completion. In this case, the agent passively receives input from users and generates the strategies to refine and clarify users' goals, as it only receives the context information directly provided by users. Please note that in multi-agent systems, an agent can send prompts by invoking the API of another agent to assign specific task, while the latter agent analyses the received information and determine the goal.

% The current description of context is too limited. Context engineering should capture everything about a user and their surroundings over time, including a complete repository of artefacts being worked on, all relevant tools used in recent sessions, and relevant conversation histories, relevant positive/negative examples. This context should be automatically appended to my prompt in passive mode, and in proactive mode, it should alert me to relevant issues with a low false positive rate, avoiding unnecessary interruptions like the infamous 'Clippy'.

\vspace{0.5em}\noindent \textbf{Consequences:} 

Benefits:
\begin{itemize}
  \item \textit{Interactivity.} Users or other agents can interact with an agent via a dialogue interface or related APIs.

  \item \textit{Goal-seeking.} The agent can analyse user-provided context and retrieve related information from memory, to identify and determine the objectives and create corresponding strategies.
  
  %create refined strategies to achieve the goals and generate accurate responses to users.

  %\item \textit{Intuitiveness.} Users can directly specify the goals, provide context to the agent, and receive the agent's responses through the dialogue interface, which is intuitive and easy to use.

  \item \textit{Efficiency.} Users can directly send prompts to the agent through the dialogue interface, which is intuitive and easy to use.

\end{itemize}

Drawbacks: 
\begin{itemize}
   \item \textit{Reasoning uncertainty.} Users may have assorted backgrounds and experiences. Unclear or ambiguous context information may intensify the reasoning uncertainties, especially considering there are no standard prompt requirements.
\end{itemize}

\vspace{0.5em}\noindent \textbf{Known uses:} 
\begin{itemize}
   \item Liu et al.~\cite{liu2023ai} designed an agent that can communicate with users and help refine research questions via a dialogue interface.

   \item Kannan et al.~\cite{kannan2023smart} proposed an agent for users to decompose and allocate tasks to robots through a dialogue interface. 

   \item \textit{HuggingGPT}\footnote{\url{https://huggingface.co/spaces/microsoft/HuggingGPT}\label{HuggingGPT}}. HuggingGPT can generate responses to address user requests via a chatbot. Users' requests including complex intents can be interpreted as their intended goals.

\end{itemize}

\vspace{0.5em}\noindent \textbf{Related patterns:} 

\begin{itemize}
    \item \textit{Proactive goal creator.} \textit{Proactive goal creator} can be regarded an alternative of \textit{passive goal creator} enabling multimodal context injection.

    \item \textit{Prompt/response optimiser}. \item \textit{Passive goal creator} can first handle users' inputs and transfer the goals and relevant context information to \textit{prompt/response optimiser} for prompt refinement.
\end{itemize}

% \begin{figure*}[t]
% 	\centering
% 	\subfigure[Passive goal creator.]{
% 	\centering
%     \includegraphics[width=0.4\columnwidth]{figures/passive_goal_creator.pdf}
%     }
%     \subfigure[Proactive goal creator.]{
%     \centering
%     \includegraphics[width=0.4\columnwidth]{figures/proactive_goal_creator.pdf}
%     }

% 	\caption{Goal creator patterns for context engineering.}
% 	\label{fig:goal_creator}
% \end{figure*}

\subsection{Proactive Goal Creator}

\vspace{0.5em}\noindent \textbf{Summary:} Proactive goal creator anticipates users' goals by understanding human interactions and capturing the context via relevant tools.

\vspace{0.5em}\noindent \textbf{Context:} Users explain the goals that the agent is expected to achieve in the prompt.

\vspace{0.5em}\noindent \textbf{Problem:} The context information collected via solely a dialogue interface may be limited, and result in inaccurate responses to users' goals.

\vspace{0.5em}\noindent \textbf{Forces:} 

\begin{itemize}
  % \item \textit{Interactivity.} Users are required to interact with the agent to provide instructions and receive responses.

  % \item \textit{Goal-seeking.} The agent requires as much information as possible to understand users' goals.

  \item \textit{Underspecification.} i) Users may not be able to provide thorough context information and specify precise goals to agents. ii) Agents can only retrieve limited information from the memory.

  \item \textit{Accessibility.} Users with specified disabilities may not be able to directly interoperate with the agent via \textit{passive goal creator}.
\end{itemize}

\begin{figure}[!ht]
    \centering
    \includegraphics[width=0.7\columnwidth]{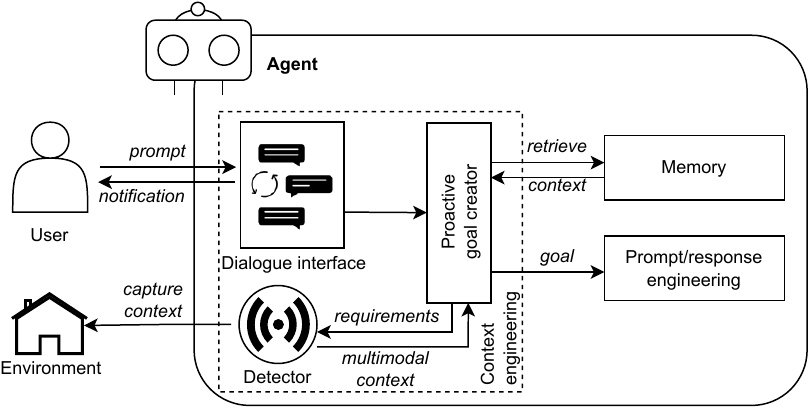}	
    \caption{Proactive goal creator.}
    \label{fig:proactive}
\end{figure}

\vspace{0.5em}\noindent \textbf{Solution:} Fig.~\ref{fig:proactive} illustrates a simple graphical representation of \textit{proactive goal creator}. In addition to the prompts received from dialogue interface, and relevant context retrieved from memory, the \textit{proactive goal creator} can anticipate users' goals by sending requirements to detectors, which will then capture and return the user's surroundings with multimodal context information for further analysis and comprehension to generate the goals, for instance, identifying the user's gestures through cameras, recognising application UI layout via screenshots, etc. Please note the \textit{proactive goal creator} should notify users about context capturing and other relevant issues with a low false positive rate, to avoid unnecessary interruptions. In addition, the captured environment information can be stored in the agent's memory (or knowledge base) to establish ``world models''~\cite{ha2018world, lecun2022path} to continuously improve its ability to comprehend the real world.

% multiple methods in addition to the explicit user text prompt. \textit{Proactive goal creator} is facilitated by virtue of analysing multimodal context information, e.g., screenshot and recording, microphone, mouse clicks, eye tracking, typing, etc.

\vspace{0.5em}\noindent \textbf{Consequences:} 

Benefits:
\begin{itemize}
  \item \textit{Interactivity.} An agent can interact with users or other agents by anticipating their decisions proactively with captured multimodal context information.

  %Users can interact with the agent which can capture multimodal information. In addition, in the multi-agent environment, \textit{proactive goal creator} can facilitate the interactions between different agents by anticipating their decisions proactively.

  \item \textit{Goal-seeking.} The multimodal input can provide more detailed information for the agent to understand users' goals, and increase the accuracy and completeness of goal achievement.

  \item \textit{Accessibility.} Additional tools can help capture the sentiments and other context information from disabled users, ensuring accessibility and broadening the human values of foundation model-based agents.

\end{itemize}

Drawbacks: 
\begin{itemize}
   \item \textit{Overhead.} i) \textit{Proactive goal creator} is enabled by the multimodal context information captured by relevant tools, which may increase the cost of the agent. ii) Limited context information may increase the communication overhead between users and agents.
\end{itemize}

\vspace{0.5em}\noindent \textbf{Known uses:} 
\begin{itemize}
   \item \textit{GestureGPT}~\cite{zeng2023gesturegpt}. GestureGPT can decipher users' hand gesture descriptions and hence comprehend users' intents.

   \item Zhao et al.~\cite{zhao2023seehow} proposed a programming screencast analysis tool that can extract the coding steps and code snippets.

   \item \textit{ProAgent}~\cite{zhang2023proagent}. ProAgent can observe the behaviours of other teammate agents, deduce their intentions, and adjust the planning accordingly.
\end{itemize}

\vspace{0.5em}\noindent \textbf{Related patterns:} 

\begin{itemize}
    \item \textit{Passive goal creator.} \textit{Proactive goal creator} can be regarded an alternative of \textit{passive goal creator} enabling multimodal context injection.

    \item \textit{Prompt/response optimiser}. \textit{Proactive goal creator} can first handle users' inputs and transfer the goals and relevant context information to \textit{prompt/response optimiser} for prompt refinement.

    \item \textit{Multimodal guardrails}. Multimodal guardrails can help process the multimodal data captured by proactive goal creator.
\end{itemize}

\subsection{Prompt/Response Optimiser}

\vspace{0.5em}\noindent \textbf{Summary:} Prompt/response optimiser refines the prompts/responses according to the desired input or output content and format.

\vspace{0.5em}\noindent \textbf{Context:} Users may struggle with writing effective prompts, especially considering the injection of comprehensive context. Similarly, it may be difficult for users to understand the agent's outputs in certain cases.

%comprehensive/relevant context should be injected automatically instead of waiting for human input. 

%After obtaining users' input context and confirming the goals, the agent needs to generate prompts or responses to users.

\vspace{0.5em}\noindent \textbf{Problem:} How to generate effective prompts and standardised responses that are aligned with users' goals or objectives?

\vspace{0.5em}\noindent \textbf{Forces:} 

\begin{itemize}
  \item \textit{Standardisation.} Prompts and responses may vary in structure, format, and content, which will lead to potential confusion or inconsistent behaviours of the agent.

  \item \textit{Goal alignment.} Ensuring that prompts and responses are aligned with the ultimate goal or objective can facilitate the agent to achieve desired results.

  \item \textit{Interoperability.} The generated prompts and responses may be directly input to other components, external tools or agents for completing further tasks.
\end{itemize}

\begin{figure}[!ht]
    \centering
    \includegraphics[width=0.45\columnwidth]{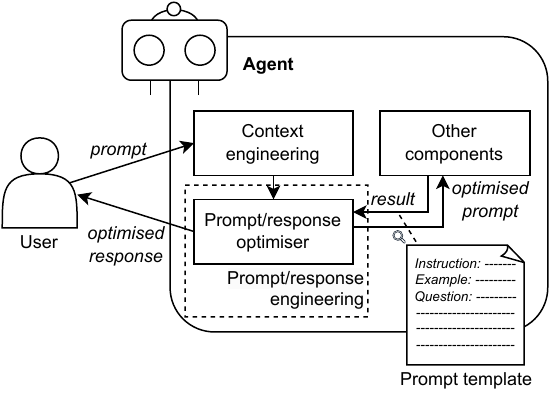}	
    \caption{Prompt/response optimiser.}
    \label{fig:response_generator}
\end{figure}

\vspace{0.5em}\noindent \textbf{Solution:} Fig.~\ref{fig:response_generator} illustrates a high-level graphical representation of \textit{prompt/response optimiser}. A user may input initial prompts to the agent, however, such prompts may be ineffective due to the lack of relevant context, unintentional injection attacks, redundancy, etc. In this regard, \textit{prompt/response optimiser} can construct refined prompts and responses adhering to predefined constraints and specifications. These constraints and specifications outline the desired content and format for the inputs and outputs, ensuring alignment with the ultimate goal. A prompt or response template is often used in the \textit{prompt/response optimiser} as a factory for creating specific instances of prompts or responses~\cite{zhao2023expel, schumann2023velma}. This template offers a structured approach to standardise the queries and responses, improving the accuracy of the responses and facilitate their interoperations with external tools or agents. For instance, a prompt template can contain the instructions to an agent, some examples for few-shot learning, and the question/goal for the agent to work.

\vspace{0.5em}\noindent \textbf{Consequences:} 

Benefits:
\begin{itemize}
  \item \textit{Standardisation.} \textit{Prompt/response optimiser} can create standardised prompts and responses regarding the requirements specified in the template.

  \item \textit{Goal alignment.} The optimised prompts and responses adhere to user-defined conditions, hence they can achieve higher accuracy and relevance to the goals.

  \item \textit{Interoperability.} Interoperability between agent and external tools is facilitated by \textit{prompt/response optimiser}, which can provide consistent and well-defined prompts and responses for task execution.

  \item \textit{Adaptability.} \textit{Prompt/response optimiser} can accommodate different constraints, specifications, or domain-specific requirements by refining the template with a knowledge base. 

\end{itemize}

Drawbacks: 
\begin{itemize}
%   \item \textit{Inflexibility.} 

   \item \textit{Underspecification.} In certain cases, it may be difficult for \textit{prompt/response optimiser} to capture and incorporate all relevant contextual information effectively, especially considering the ambiguity of users' input, and dependency on context engineering. Consequently, the optimiser may struggle to generate appropriate prompts or responses.

   \item \textit{Maintenance overhead.} Updating and maintaining prompt or response templates may cause significant overhead. Changes in requirements may necessitate modifying multiple templates, which is time-consuming and error-prone.
\end{itemize}

\vspace{0.5em}\noindent \textbf{Known uses:} 
\begin{itemize}
   \item \textit{LangChain}\footnote{\url{https://api.python.langchain.com/en/latest/prompts/langchain_core.prompts.prompt.PromptTemplate.html}}. LangChain provides prompt templates for practitioners to develop custom foundation model-based agents.

   \item \textit{Amazon Bedrock}\footnote{\url{https://docs.aws.amazon.com/bedrock/latest/userguide/advanced-prompts-configure.html}\label{Bedrock}}. Users can configure prompt templates in Amazon Bedrock, defining how the agent should evaluate and use the prompts. 

   \item \textit{Dialogflow}\footnote{\url{https://cloud.google.com/dialogflow}}. Dialogflow allows users to create generators to specify agent behaviours and responses at runtime.

\end{itemize}

\vspace{0.5em}\noindent \textbf{Related patterns:} 

\begin{itemize}
    \item \textit{Passive goal creator} and \textit{proactive goal creator} can first handle users' inputs and transfer the goals and relevant context information to \textit{prompt/response optimiser} for prompt refinement.

    \item \textit{Self-reflection, cross-reflection,} and \textit{human-reflection.} The reflection patterns can be applied to assess and refine the output of \textit{prompt/response optimiser}.

    \item \textit{Agent adapter.} \textit{Prompt/response optimiser} can improve users' inputs, and the optimised prompts can be sent to other agents for goal achievement, while \textit{agent adapter} focuses more on the utilisation of external tools.
\end{itemize}

\subsection{Retrieval Augmented Generation (RAG)} 

\vspace{0.5em}\noindent \textbf{Summary:} Retrieval augmented generation techniques enhance the knowledge updatability of agents for goal achievement, and maintain data privacy of on-premise foundation model-based agents/systems implementations.

\vspace{0.5em}\noindent \textbf{Context:} Large foundational model-based agents are not equipped with knowledge related to explicitly specific domains, especially on highly confidential and privacy-sensitive local data, unless they are fine-tuned for pre-trained using domain data.

\vspace{0.5em}\noindent \textbf{Problem:} Given a task, how can agents conduct reasoning with data/knowledge that are not learned by the foundation models through model training?

\vspace{0.5em}\noindent \textbf{Forces:} 

\begin{itemize}
  \item \textit{Lack of knowledge.} The reasoning process may be unreliable when the agent is required to accomplish domain-specific tasks that the agent has no such knowledge reserve.
  %It is unrealistic to build an almighty agent which acquires various knowledge,

  \item \textit{Overhead.} Fine-tuning large foundation model using local data or training a large foundation model locally consumes high amount of computation and resource costs.
  
  \item \textit{Data Privacy.} Local data are confidential to be used to train or fine-tune the models.
\end{itemize}

\begin{figure}[!ht]
    \centering
    \includegraphics[width=0.5\columnwidth]{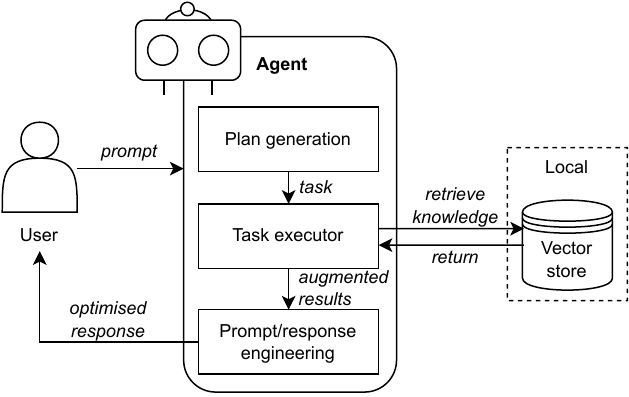}	
    \caption{Retrieval augmented generation.}
    \label{fig:RAG}
\end{figure}

\vspace{0.5em}\noindent \textbf{Solution:} Fig.~\ref{fig:RAG} illustrates a high-level graphical representation of \textit{retrieval augmented generation}. RAG is a technique for enhancing the accuracy and reliability of agents with facts retrieved from other sources (internal or online data). The knowledge gaps where the agents are lacking in memory are filled using the parameterized knowledge generated in vector databases. For instance, after plan generation, specific tasks may require information that is not within the original agent memory. The agent can hence retrieve information from the parameterised knowledge and use for task completion, while the augmented responses will be returned back to the user after optimisation. In particular, the implementation of RAG consists of the following steps: i) Determine the data sources. ii) Define the data structure for indexing raw data (i.e., text, image, video and audio) into embeddings or knowledge graphs. iii) Given a certain query, task executor encodes this query and searches the knowledge base (e.g., vector store) to retrieve the most relevant information. iv) Task executor processes the obtained data through reranking and filtering to produce a more informed and accurate response.

The retrieval process requires zero pretraining or fine-tuning of the model served by the agent which preserves the data privacy of local data, reduces training and computation costs, and also provides up-to-date and more precise information required. The retrieved local data can be sent back to the agent via prompts (need to consider the context window size), whereas the agent is able to process the information and generate plans via in-context learning. Currently there is a cluster of RAG techniques focusing on various enhancement aspects, data sources and applications~\cite{hu2024rag}, for instance, federated RAG~\cite{wang2024feb4rag}, graph RAG~\cite{graph_rag}, etc. Further, Retrieval Interleaved Generation can be considered a related technique of RAG where the agent can dynamically access external knowledge throughout the response generation phase.

%When a prompt requests information that is not within the original memory that the agent contains, it can retrieve information from the parameterized knowledge and provide augemented response to the user. The retrieval process requires zero pretraining or fine-tuning of the model served by the agent which preserves the data privacy of local data, reduces training and computation costs, and also provides up-to-date and more precise information required.

\vspace{0.5em}\noindent \textbf{Consequences:} 

Benefits:
\begin{itemize}
  \item \textit{Knowledge retrieval.} Agents can search and retrieve knowledge related to the given tasks, which ensures the reliability of reasoning steps.

  \item \textit{Updatability.} The prompts/responses generated using RAG by the agent on internal or online data are updatable by the complimentary parameterized knowledge.

  \item \textit{Data privacy.} The agent can retrieve additional knowledge from local datastores, which ensures data privacy and security.

  \item \textit{Cost-efficiency.} Under the data privacy constraint, RAG can provide essential knowledge to the agent without training a new foundation model entirely. This reduced the training costs.

\end{itemize}

Drawbacks: 
\begin{itemize}
   \item \textit{Maintenance overhead.} Maintenance and update of the parameterized knowledge in the vector store requires additional computation and storage costs.
   
   \item \textit{Data limitation.} The agents still mainly rely on the data it has been trained on to generate prompts. This can impact the quality and accuracy of the generated content in those specific domains.
\end{itemize}

\vspace{0.5em}\noindent \textbf{Known uses:} 
\begin{itemize}
   %\item \textit{Databricks LLMs for Customer Service and Support\footnote{\url{https://www.databricks.com/solutions/accelerators/llms-customer-service-and-support}}.} Databricks provide service that uses pretrained foundation models on the Databricks Lakehouse where organizations can ingest enterprise data from various knowledge bases to build a context-enabled FM-based chatbot.
   
   %\item \textit{Azure AI Search\footnote{\url{https://learn.microsoft.com/en-us/azure/search/retrieval-augmented-generation-overview}}.} Microsoft's Azure AI Search in RAG solutions includes Azure AI Studio which uses a vector index and retrieval augmentation, Azure OpenAI Studio which uses a search index with or without vectors, and Azure Machine Learning, which uses a search index as a vector store in a prompt flow.

   %\item \textit{Weaviate\footnote{\url{https://weaviate.io/}}.} Weaviate is an open-source vector database that supports the retrieval augmented generation. It has been integrated with generative AI solutions such as OpenAI, LangChain, Google, etc.

   \item \textit{LinkedIn}\footnote{\url{https://www.linkedin.com/blog/engineering/generative-ai/musings-on-building-a-generative-ai-product}}. LinkedIn applies RAG to construct the pipeline of foundation model based agents, which can search appropriate case studies to respond users.

   %https://arxiv.org/abs/2404.19543
   %Corrective retrieval augmented generation
   %Learning to filter context for retrieval-augmented generation
   %Retrieval-augmented generation to improve math question-answering: Trade-offs between groundedness and human preference

   \item Yan et al.~\cite{yan2024corrective} devise a retrieval evaluator which can output a confidence degree after assessing the quality of retrieved data. The solution can improve the robustness and overall performance of RAG for agents.

   \item Levonian et al.~\cite{levonian2023retrieval} apply RAG with GPT-3.5, developing an agent that can retrieve the contents of a high-quality open-source math textbook to generate responses to students.

\end{itemize}

\vspace{0.5em}\noindent \textbf{Related patterns:} \textit{Retrieval augmented generation} can complement all other patterns by providing extra context information from the local datastore.

\subsection{One-Shot Model Querying}
%\boming{not sure about this one. There are systems/agents generating many solutions and then filter and score them and output the selected one(s). See examples like AlphaCode 2 and MedPrompt in \url{https://bair.berkeley.edu/blog/2024/02/18/compound-ai-systems/}. In addition, MedPrompt ``Answers medical questions by searching for similar examples to construct \textbf{a few-shot prompt}, adding model-generated chain-of-thought for each example, and generating and judging up to 11 solutions".}

\vspace{0.5em}\noindent \textbf{Summary:} The foundation model is accessed in a single instance to generate all necessary steps for the plan.

\vspace{0.5em}\noindent \textbf{Context:} When users interact with the agent for specific goals, the included foundation model is queried for plan generation.

\vspace{0.5em}\noindent \textbf{Problem:} How can the agent generate the steps for a plan efficiently?

\vspace{0.5em}\noindent \textbf{Forces:} 

\begin{itemize}
  \item \textit{Efficiency.} For certain pressing tasks, the agent should be able to conduct planning and respond in a short amount of time.

  \item \textit{Overhead.} Users need to pay for each interaction with commercial foundation models.
\end{itemize}

\begin{figure}[!ht]
    \centering
    \includegraphics[width=0.35\columnwidth]{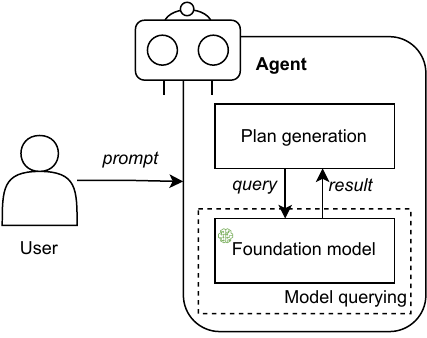}	
    \caption{One-shot model querying.}
    \label{fig:one_shot}
\end{figure}

\vspace{0.5em}\noindent \textbf{Solution:} Fig.~\ref{fig:one_shot} illustrates interactions between user and agent within \textit{one-shot model query}. In this scenario, the agent queries the incorporated foundation model to generate a corresponding plan based on user-specified goals and constraints. The foundation model is only queried for once in regard to the user's requirements (e.g. limited budget), to comprehend the provided inputs. In this manner, the agent can devise a multi-step plan to achieve a broad goal, and provide a holistic explanation for this plan without delving into detailed reasoning steps. Please note that this pattern is applicable when other components query the integrated foundation model.

%After a user specifies goals and constraints in one prompt, the agent will query the incorporated foundation model to generate a corresponding response (e.g. plan for action). The foundation model does not require multiple interactions to comprehend the context and requirements. In this manner, the agent can devise a multi-step plan to achieve a broad goal, and provide a holistic explanation for this plan without delving into detailed reasoning steps.

%The generated plan should be confirmed by the user before the enclosed tasks are assigned to other agents or tools. %Finally, the task results are returned to the original agent for response generation.

\vspace{0.5em}\noindent \textbf{Consequences:} 

Benefits:
\begin{itemize}
   \item \textit{Efficiency.} The agent can generate a plan to achieve users' goals by querying the underlying foundation model only once, which saves consumed time.

  \item \textit{Cost-efficiency.} Users' expenses can be reduced since the foundation model is queried for one time.

  \item \textit{Simplicity.} \textit{One-shot model querying} can satisfy the tasks that do not require complex action plans.

\end{itemize}

Drawbacks: 
\begin{itemize}
   \item \textit{Oversimplification.} For complex tasks, \textit{one-shot model querying} may not be able to fully capture all requirements at one time, hence oversimplifying the tasks and cannot return a correct response.

   \item \textit{Lack of explainability.} \textit{One-shot model querying} may suffer the lack of explainability as the incorporated foundation model is queried only once, which may not provide detailed reasoning steps for plan generation.
   
   \item \textit{Size of the context window.} The response quality may be constrained considering the foundation models' current capability of handling long conversational contexts and the token limits.
\end{itemize}

\vspace{0.5em}\noindent \textbf{Known uses:} \textit{One-shot model querying} can be considered configuration or use by default when a user is leveraging a foundation model, while CoT and Zero-shot-CoT both exemplify this pattern~\cite{wang2024survey, wang-etal-2023-plan}.
% \begin{itemize}
%   \item \textit{Zero-shot prompting}~\cite{wei2022finetuned}. Foundation models have demonstrated zero-shot learning abilities to users' prompts, where users only need to send the questions for one time and obtain answers.
  
%   \item \textit{GestureGPT}~\cite{zeng2023gesturegpt}. GestureGPT supports zero-shot gesture understanding and provides a candidate function list regarding the current gesture context with \textit{one-shot model querying}.
  
%   \item \textit{Zero-shot-CoT}~\cite{NEURIPS2022_8bb0d291}. Zero-shot-CoT can generate a plausible reasoning path by querying the foundation model for once, achieving outperformance over standard zero-shot prompting approaches.

%   %Large Language Models are Zero-Shot Reasoners
% \end{itemize}

\vspace{0.5em}\noindent \textbf{Related patterns:} 

\begin{itemize}
    \item \textit{Incremental model querying.} \textit{Incremental model querying} can be regarded an alternative of \textit{one-shot model querying} with iteration. 
    
    \item \textit{Single-path plan generator.} \textit{One-shot model querying} enables the generation of single-path plans by only querying the foundation model for one time.

    \item \textit{Multimodal guardrails}. Multimodal guardrails serve as an intermediate layer, managing the inputs and outputs of model querying.
\end{itemize}

% \begin{figure*}[t]
% 	\centering
% 	\subfigure[One-shot model querying.]{
% 	\centering
%     \includegraphics[width=0.4\columnwidth]{figures/one-shot.pdf}
%     }
%     \subfigure[Incremental model querying.]{
%     \centering
%     \includegraphics[width=0.4\columnwidth]{figures/incremental.pdf}
%     }

% 	\caption{Model querying patterns for planning.}
% 	\label{fig:model_querying}
% \end{figure*}

\subsection{Incremental Model Querying}
%\boming{Random thought, another potential pattern: AI gateways/routers, from \url{https://bair.berkeley.edu/blog/2024/02/18/compound-ai-systems/}}

\vspace{0.5em}\noindent \textbf{Summary:} Incremental model querying involves accessing the foundation model at each step of the plan generation process.

%\liming{EcoAssistant: Using LLM Assistant More Affordably and Accurately}

\vspace{0.5em}\noindent \textbf{Context:} When users interact with the agent for specific goals, the included foundation model is queried for plan generation.
%To support iterative coding, it allows the conversational LLM as an assistant agent to converse with an automatic code executor and iteratively refine code to make the correct API calls.

\vspace{0.5em}\noindent \textbf{Problem:} The foundation model may struggle to generate the correct response at the first attempt. How can the agent conduct an accurate reasoning process?

%For example, to re-iterate, the task of code-driven question answering is both challenging and expensive. LLMs struggle to generate the correct code at the first attempt to utilize APIs, and handling a high volume of user queries using LLM services with a fee can be cost-intensive.

\vspace{0.5em}\noindent \textbf{Forces:} 

\begin{itemize}
  \item \textit{Size of the context window.} The context window of a foundation model may be limited, hence users may not be able to provide a complete and comprehensive prompt.

  \item \textit{Oversimplification.} The reasoning process may be oversimplified and hence endure uncertainties with only one attempt of model querying.
  %The generated response may suffer hallucinations of foundation models at the first attempts of querying.

  \item \textit{Lack of explainability.} The generated responses of foundation models require detailed reasoning process to preserve explainability and eventual trustworthiness.
\end{itemize}

\begin{figure}[!ht]
    \centering
    \includegraphics[width=0.35\columnwidth]{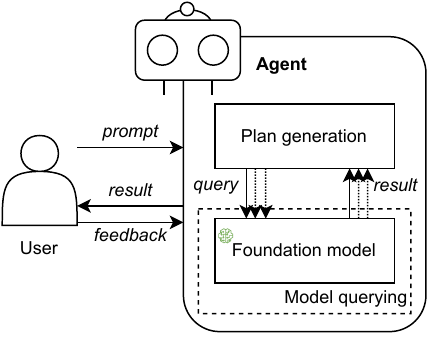}	
    \caption{Incremental model querying.}
    \label{fig:incremental}
\end{figure}

\vspace{0.5em}\noindent \textbf{Solution:} Fig.~\ref{fig:incremental} illustrates interactions between the plan generation component and integrated foundation model with \textit{incremental model querying}. The agent could engage in a step-by-step reasoning process to develop the plan for goal achievement with multiple queries to the foundation model. Meanwhile, human feedback can be provided at any time to both the reasoning process and generated plan, and adjustments can be made accordingly during model querying. The number of queries can be predefined in agent configuration or determined in user prompts. Please note that \textit{incremental model querying} can rely on a reusable template, which guides the process through context injection or an explicit workflow/plan repository and management system. This pattern is applicable when other components query the integrated foundation model.

\vspace{0.5em}\noindent \textbf{Consequences:} 

Benefits:
\begin{itemize}
  \item \textit{Supplementary context.} \textit{Incremental model querying} allows users to split the context in multiple prompts to address the issue of limited context window.

  \item \textit{Reasoning certainty.} Foundation models will iteratively refine the reasoning steps by self-checking or feedback from users.

  \item \textit{Explainability.} Users can query the foundation model to provide detailed reasoning steps through \textit{incremental model querying}.

\end{itemize}

Drawbacks: 
\begin{itemize}
   \item \textit{Overhead.} i) \textit{Incremental model querying} requires multiple interactions with the foundation model, which may increase the time consumption for planning determination. ii) The high volume of user queries may be cost-intensive when utilising commercial foundation models.
   %Such a high volume of user queries can be expensive for those who aim to develop a system using online LLM services with a fee to process these queries.
\end{itemize}

\vspace{0.5em}\noindent \textbf{Known uses:} 
\begin{itemize}
   \item \textit{HuggingGPT}\textsuperscript{\ref{HuggingGPT}}. The underlying foundation model of HuggingGPT is queried multiple times to decompose users' requests into fine-grained tasks, and then determine the dependencies and execution orders of tasks~\cite{NEURIPS2023_77c33e6a}.
   
   \item \textit{EcoAssistant}~\cite{zhang2023ecoassistant}. EcoAssistant applies a code executor interacting with the foundation model to iteratively refine code.

   \item \textit{ReWOO}~\cite{xu2023rewoo}. ReWOO queries the foundation model to i) generate a list of interdependent plans, and; ii) combine the observation evidence fetched from tools with the corresponding task.
\end{itemize}

\vspace{0.5em}\noindent \textbf{Related patterns:} 

\begin{itemize}
    \item \textit{One-shot model querying.} \textit{Incremental model querying} can be regarded an alternative of \textit{one-shot model querying} with iteration. 
    
    \item \textit{Multi-path plan generator.} The agent can capture users' preferences at each step and generate multi-path plans by iteratively querying the foundation model.

    \item \textit{Self-reflection.} \textit{Self-reflection} requires agents to query their incorporated foundation model multiple times for response review and evaluation.

    \item \textit{Human-reflection.} \textit{Human-reflection} is enabled by \textit{incremental model querying} for iterative communication between users/experts and the agent.

    \item \textit{Multimodal guardrails}. Multimodal guardrails serve as an intermediate layer, managing the inputs and outputs of model querying.
\end{itemize}

\subsection{Single-Path Plan Generator} 

\vspace{0.5em}\noindent \textbf{Summary:} Single-path plan generator orchestrates the generation of intermediate steps leading to the achievement of the user’s goal.

\vspace{0.5em}\noindent \textbf{Context:} A agent is considered ``black box'' to users, while users may care about the process of how an agent achieve users' goals.

%An agent may not be proficient at performing all tasks, while users also care about the process of how an agent produces the outputs.

%An agent needs to formulate strategies and make a plan to provide a linear, coherent path to achieve users' goals.

\vspace{0.5em}\noindent \textbf{Problem:} How can an agent efficiently formulate the strategies to achieve users' goals?

%When presented with a high-level goal or task, how can an agent effectively accomplish the goal?

\vspace{0.5em}\noindent \textbf{Forces:} 

\begin{itemize}
  \item \textit{Underspecification.} Users may assign tasks with high-level abstraction, which may be challenging for agents to handle the uncertainty or ambiguity in the provided context.

  \item \textit{Coherence.} Users and other interacting tools/agents will expect coherent responses or guidelines for achieving certain goals.

  \item \textit{Efficiency.} Uncertain decisions may affect the efficiency of an agent, which will result in reduced user satisfaction.
\end{itemize}

\begin{figure*}[!ht]
    \centering
    \includegraphics[width=0.8\columnwidth]{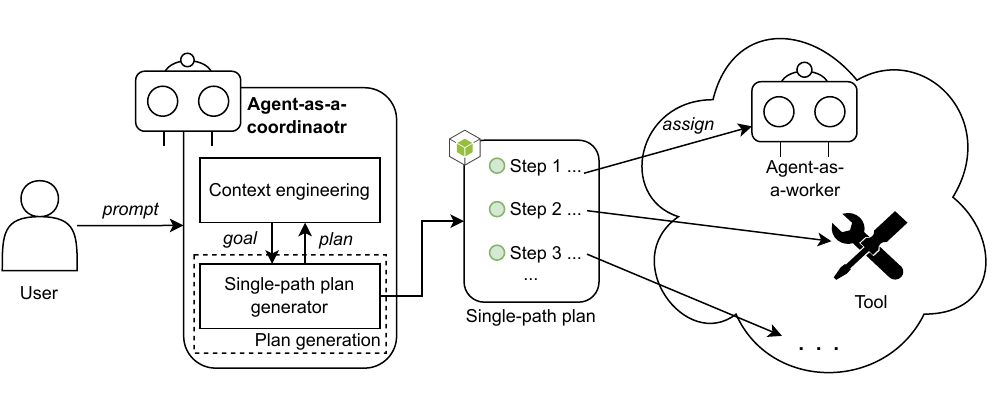}	
    \caption{Single-path plan generator.}
    \label{fig:single_path}
\end{figure*}

\vspace{0.5em}\noindent \textbf{Solution:} Fig.~\ref{fig:single_path} illustrates a high-level graphical representation of \textit{single-path plan generator}. After receiving and comprehending users' goals, the \textit{single-path plan generator} can coordinate the creation of a plan for other agents or tools and prioritise the tasks, to progressively lead towards goal accomplishment. Specifically, the plan generation process requires inference and reasoning that whether intermediate steps are actionable and optimal. Each step in this process is designed to have only a single subsequent step, forming a linear and direct plan, such as Chain-of-Thought (CoT)~\cite{wei2022chain}. Self-consistency is employed to confirm with the foundation model several times and select the most consistent answer as the final decision~\cite{wang2023rcagent}. Please note that the generated plan may have different granularity based on the given goal that complex plans may incorporate multiple workflows, processes, tasks and fine-grained steps.

\vspace{0.5em}\noindent \textbf{Consequences:} 

Benefits:
\begin{itemize}
  \item \textit{Reasoning certainty.} \textit{Single-path plan generator} generates a multi-step plan, which can reflect the reasoning process and mitigate the uncertainty or ambiguity for achieving users' goals.

  \item \textit{Coherence.} The interacting users, agents and tools are provided a clear and coherent path towards the ultimate goals.

  \item \textit{Efficiency.} \textit{Single-path plan generator} can increase efficiency in agents via pruning unnecessary steps or distractions.

\end{itemize}

Drawbacks: 
\begin{itemize}
   \item \textit{Flexibility.} A single-path plan may result in limited flexibility to accommodate diverse user preferences or application scenarios, hence users cannot customise their solutions.

   \item \textit{Oversimplification.} The agent may oversimplify the generated plan which requires multi-faceted approaches.
\end{itemize}

\vspace{0.5em}\noindent \textbf{Known uses:} 
\begin{itemize}
%   \item \textit{Amazon Bedrock}\footnote{\url{https://aws.amazon.com/blogs/aws/agents-for-amazon-bedrock-is-now-available-with-improved-control-of-orchestration-and-visibility-into-reasoning/}}. Users can step through the reasoning process and orchestration plan of agents in Bedrock.
   
   \item \textit{LlamaIndex}\footnote{\url{https://docs.llamaindex.ai/en/stable/examples/finetuning/react_agent/react_agent_finetune/}}. LlamaIndex fine-tunes a ReAct Agent to achieve better performance with \textit{single-path plan generator} via CoT.

   \item \textit{ThinkGPT}\footnote{\url{https://github.com/jina-ai/thinkgpt}}. ThinkGPT provides a toolkit to facilitate the implementation of \textit{single-path plan generator} pattern.

   \item Zhang et al.\cite{zhang2023igniting} promote the implementation by elucidating the basic mechanisms and paradigm shift of CoT.
  
\end{itemize}

\vspace{0.5em}\noindent \textbf{Related patterns:} 

\begin{itemize}
    \item \textit{One-shot model querying.} \textit{One-shot model querying} enables the generation of single-path plans by only querying the foundation model for one time.

    \item \textit{Multi-path plan generator.} \textit{Multi-path plan generator} can be regarded an alternative of \textit{single-path plan generator} for customised strategy.

    \item \textit{Self-reflection.} \textit{Single-path plan generator} and \textit{self-reflection} both contribute to self-Consistency with Chain of Thought.
\end{itemize}

\subsection{Multi-Path Plan Generator}

\vspace{0.5em}\noindent \textbf{Summary:} Multi-path plan generator allows for creating multiple choices at each intermediate step leading to achieving users' goals.

\vspace{0.5em}\noindent \textbf{Context:} A agent is considered ``black box'' to users, while users may care about the process of how an agent achieve users' goals.
%An agent may not be proficient at performing particular tasks, while users also care about the process of how an agent produces the outputs.

%An agent needs to formulate strategies and make a plan to provide a linear, coherent path to achieve users' goals.

\vspace{0.5em}\noindent \textbf{Problem:} How can an agent generate a high-quality, coherent, and efficient solution considering inclusiveness and diversity when presented with a complex task or problem?

%How can agents generate a high-quality, coherent, and efficient solution when presented with a complex task or problem?

\vspace{0.5em}\noindent \textbf{Forces:} 

\begin{itemize}
  \item \textit{Underspecification.} Users may assign tasks with high-level abstraction, which may be challenging for agents to handle the uncertainty or ambiguity in the provided context.

  \item \textit{Coherence.} Users and other interacting tools/agents will expect coherent responses or guidelines for achieving certain goals.

  \item \textit{Alignment to human preference.} Certain goals require agents to capture users' preferences, to provide customised solutions.

  \item \textit{Oversimplification.} For particular complex tasks, agents may oversimplify the reasoning process, hence the provided solutions cannot satisfy users' requirements.
\end{itemize}

\begin{figure*}[!ht]
    \centering
    \includegraphics[width=0.8\columnwidth]{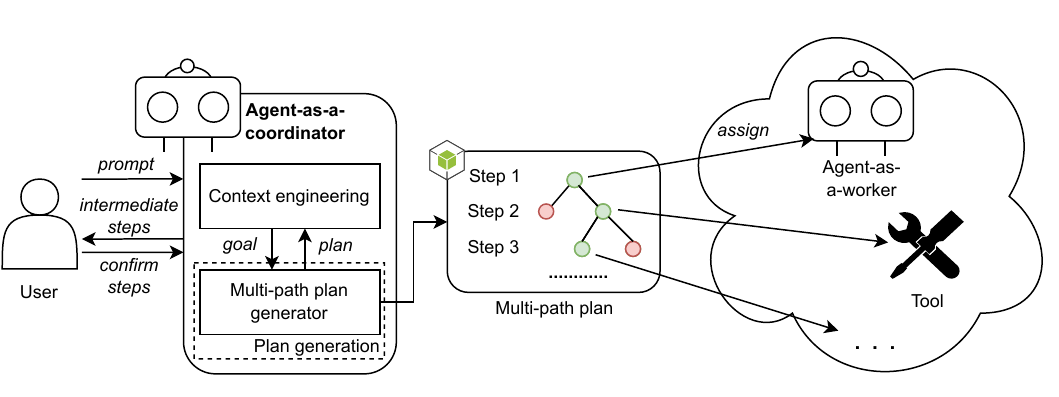}	
    \caption{Multi-path plan generator.}
    \label{fig:multiple_path}
\end{figure*}

\vspace{0.5em}\noindent \textbf{Solution:} Fig.~\ref{fig:multiple_path} illustrates a high-level graphical representation of \textit{multi-path plan generator}. Based on \textit{single-path plan generator}, \textit{multi-path plan generator} can create multiple choices at each step towards the achievement of goals, which requires the underlying foundation model to tease out the eligible and actionable activities for each choice in the previous step. Specifically, users' preferences may influence the subsequent intermediate steps, leading to different eventual plans. The employment of involved agents and tools will be adjusted accordingly. Tree-of-Thoughts~\cite{yao2024tree} exemplifies this design pattern.

\vspace{0.5em}\noindent \textbf{Consequences:} 

Benefits:
\begin{itemize}
   \item \textit{Reasoning certainty.} \textit{Multi-path plan generator} can generate a plan with multiple choices of intermediate steps to resolve the uncertainty or ambiguity within reasoning process.

  \item \textit{Coherence.} The interacting users, agents and tools are provided a clear and coherent path towards the ultimate goals.

  \item \textit{Alignment to human preference.} Users can confirm each intermediate step to finalise the planning, hence human preferences are absorbed in the generated customised strategy.

  \item \textit{Inclusiveness.} The agent can specify multiple directions in the reasoning process for complex tasks.

\end{itemize}

Drawbacks: 
\begin{itemize}
   \item \textit{Overhead.} Task decomposition and multi-plan generation may increase the communication overhead between the user and agent.
\end{itemize}

\vspace{0.5em}\noindent \textbf{Known uses:} 
\begin{itemize}

   \item \textit{AutoGPT}\textsuperscript{\ref{AutoGPT}}. AutoGPT can make informed decisions by incorporating Tree-of-Thoughts as the \textit{multi-path plan generator}.
   
   \item \textit{Gemini}\footnote{\url{https://www.youtube.com/watch?v=v5tRc_5-8G4}}. For a task, Gemini can generate multiple choices for users to decide. Upon receiving users' responses, Gemini will provide multiple choices for the next step.

   \item \textit{Open AI}\footnote{\url{https://github.com/princeton-nlp/tree-of-thought-llm}}. GPT-4 was leveraged to implement a \textit{multi-path plan generator} based on Tree-of-Thoughts.

\end{itemize}

\vspace{0.5em}\noindent \textbf{Related patterns:} 

\begin{itemize}
    \item \textit{Incremental model querying.} The agent can capture users' preferences at each step and generate multi-path plans by iteratively querying the foundation model.
    
    \item \textit{Single-path plan generator.} \textit{Multi-path plan generator} can be regarded an alternative of \textit{single-path plan generator} for customised strategy.

    \item \textit{Human-reflection.} \textit{Multi-plan generator} creates plans with various directions, and \textit{human-reflection} can help finalise the plan with user feedback to determine the choice of each intermediate step.
\end{itemize}

\subsection{Self-Reflection}

\vspace{0.5em}\noindent \textbf{Summary:} Self-reflection enables the agent to generate feedback on the plan and reasoning process and provide refinement guidance from themselves.

%\liming{Reflexion: Language Agents with Verbal Reinforcement Learning}

\vspace{0.5em}\noindent \textbf{Context:} Given users' goals and requirements, the agent will generate a plan to decompose the goals into a set of tasks for achieving the goals.

\vspace{0.5em}\noindent \textbf{Problem:} A generated plan may be affected by hallucinations of the foundation model, how to review the plan and reasoning steps and incorporate feedback efficiently?
%the agent needs to incorporate feedback to refine the plan.

\vspace{0.5em}\noindent \textbf{Forces:} 

\begin{itemize}
  \item \textit{Reasoning uncertainty.} There may be inconsistencies or uncertainties embedded in the agent's reasoning process, affecting the task success rate and response accuracy.

  \item \textit{Lack of explainability.} The trustworthiness of the agent can be disturbed by the issue of transparency and explainability of how the plan is generated.

  \item \textit{Efficiency.} Certain goals require the plan to be finalised within a specific time period.
\end{itemize}

\begin{figure*}[t]
    \centering
    \includegraphics[width=0.82\columnwidth]{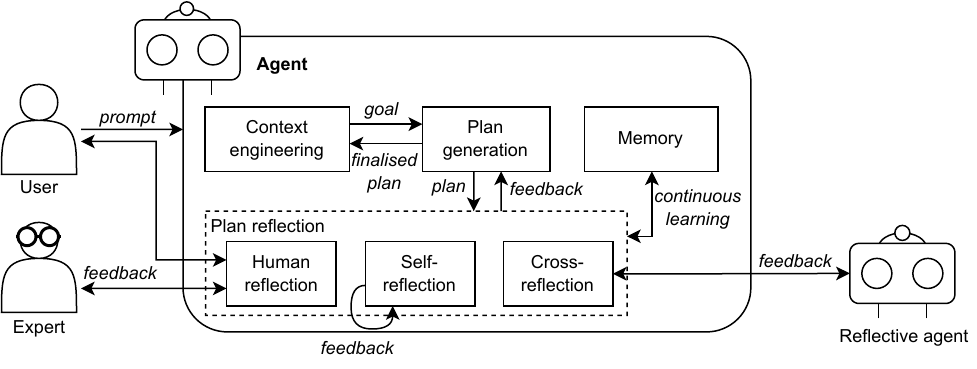}	
    \caption{Plan reflection pattern.}
    \label{fig:reflection}
\end{figure*}

\vspace{0.5em}\noindent \textbf{Solution:} Fig.~\ref{fig:reflection} depicts a high-level graphical representation of \textit{self-reflection}. In particular, reflection is an optimisation process formalised to iteratively review and refine the reasoning process and generated contents of the agent. The user prompts specific goals to the agent, which then generates a plan to accomplish users' requirements. Subsequently, the user can instruct the agent to reflect on the plan and the corresponding reasoning process. The agent will backtrack the inference process to verify whether certain intermediate results are incorrect and hence misleading all subsequent steps, then adjust and align its reasoning process to create a refined plan accordingly. Such reflection processes and results can be saved in the agent's memory for continuous learning. The finalised plan will be carried out step by step. Self-consistency~\cite{huang2022large} exemplifies this pattern.

%The Reflexion process Reflexion is formalized as an iterative optimization process in 1. In the first trial, the Actor produces a trajectory τ0 by interacting with the environment. The Evaluator then produces a score r0 which is computed as rt = Me(τ0). rt is only a scalar reward for trial t that improves as task-specific performance increases. After the first trial, to amplify r0 to a feedback form that can be used for improvement by an LLM, the Self-Reflection model analyzes the set of {τ0 , r0 } to produce a summary sr0 which is stored in the memory mem. srt is a verbal experience feedback for trial t. The Actor, Evaluator, and Self-Reflection models work together through trials in a loop until the Evaluator deems τt to be correct. As mentioned in 3, the memory component of Reflexion is crucial to its effectiveness. After each trial t, srt, is appended mem. In practice, we bound mem by a maximum number of stored experiences, Ω (usually set to 1-3) to adhere to max context LLM limitation.

\vspace{0.5em}\noindent \textbf{Consequences:} 

Benefits:
\begin{itemize}
    \item \textit{Reasoning certainty.} Agents can evaluate their own responses and reasoning procedure to check whether there are any errors or inappropriate outputs, and make refinement accordingly.

    \item \textit{Explainability.} \textit{Self-reflection} allows the agent to review and explain its reasoning process to users, facilitating better comprehension of the agent's decision-making process.

    \item \textit{Continuous improvement.} The agent can continuously update the memory or knowledge base and the manner of formalising the prompts and knowledge, to provide more reliable and coherent output to users without or with fewer reflection steps.

    \item \textit{Efficiency.} On one hand, it is time-saving for the agent to self-evaluate its response, as no additional communication overhead is cost compared to other reflection patterns. On the other hand, the agent can provide more accurate responses in the future to reduce the overall reasoning time consumption considering the continuous improvement.

  % \item \textit{Understandability.} 
  %   In the case of tool-usage that may be too hard for humans to understand, self-reflections could be monitored to ensure proper intent before using the tool.
  %   \item \textit{Hallucination.} 
  %   Self-reflection could use to mitigate hallucination in LLM.
    
\end{itemize}

Drawbacks: 
\begin{itemize}
   \item \textit{Reasoning uncertainty.} The evaluation result is dependent on the complexity of \textit{self-reflection} and the agent's competence in assessing its generated responses.

   \item \textit{Overhead.} i) \textit{Self-reflection} can increase the complexity of an agent, which may affect the overall performance. ii) Refining and maintaining agents with self-reflection capabilities requires specialised expertise and development process.
\end{itemize}

\vspace{0.5em}\noindent \textbf{Known uses:} 
\begin{itemize}
   \item \textit{Reflexion}~\cite{NEURIPS2023_1b44b878}. Reflexion employs a \textit{self-reflection} model which can generate nuanced and concrete feedback based on the success status, current trajectory, and persistent memory.

   \item \textit{Bidder agent}~\cite{chen2023put}. A replanning module in this agent utilises \textit{self-reflection} to create new textual plans based on the auction’s status and new context information.

   \item \textit{Generative agents}~\cite{Generative_Agents}. Agents perform reflection two or three times a day, by first determining the objective of reflection according to the recent activities, then generating a reflection which will be stored in the memory stream.

   %https://twitter.com/DYtweetshere/status/1631349179934203904
\end{itemize}

\vspace{0.5em}\noindent \textbf{Related patterns:} 

\begin{itemize}
    \item \textit{Prompt/response optimiser.} \textit{Self-reflection} can be applied to assess and refine the output of \textit{prompt/response optimiser}.

    \item \textit{Incremental model query.} \textit{Self-reflection} requires agents to query their incorporated foundation model multiple times for response review and evaluation.

    \item \textit{Single-path plan generator.} \textit{Single-path plan generator} and \textit{self-reflection} both contribute to self-Consistency with Chain of Thought.
\end{itemize}

\subsection{Cross-Reflection}

\vspace{0.5em}\noindent \textbf{Summary:} Cross-reflection uses different agents or foundation models to provide feedback and refine the generated plan and corresponding reasoning procedure.

\vspace{0.5em}\noindent \textbf{Context:} The agent generates a plan to achieve users' goals, while the quality of this devised plan should be assessed.

\vspace{0.5em}\noindent \textbf{Problem:} When an agent has limited capability and cannot conduct reflection with satisfying performance, how to evaluate the output and reasoning steps of this agent?

\vspace{0.5em}\noindent \textbf{Forces:} 

\begin{itemize}
  \item \textit{Reasoning uncertainty.} The inconsistencies and errors in the agent's reasoning process may reduce response accuracy and affect the overall trustworthiness.

  \item \textit{Lack of explainability.} The trustworthiness of the agent can be disturbed by the issue of transparency and explainability of how the plan is generated.

  \item \textit{Limited capability.} An agent may not be able to perform reflection well due to its limited capability and the complexity of self-reflection.
\end{itemize}

\vspace{0.5em}\noindent \textbf{Solution:} Fig.~\ref{fig:reflection} includes a high-level graphical representation of \textit{cross-reflection}. If an agent cannot generate accurate results or precise planning steps via reflecting its outputs, users can prompt the agent to query another agent which is specialised in reflection. The latter agent can review and evaluate the logged outputs and relevant reasoning steps of the original agent, and provide refinement suggestions. This process can be iterative until the reflective agent confirms the plan. In addition, multiple agents can be queried for reflection to generate comprehensive responses.

\vspace{0.5em}\noindent \textbf{Consequences:} 

Benefits:
\begin{itemize}
  \item \textit{Reasoning certainty.} The agent's outputs and respective methodology are assessed and refined by other agents to ensure the reasoning certainty and response accuracy.

  \item \textit{Explainability.} Multiple agents can be employed to review the reasoning process of the original agent, providing thorough explanations to the user.

%  \item \textit{Interoperability.} The original agent needs to interact with other agents for reflection, while other agents may also interact with each other to finalise the feedback.

  \item \textit{Inclusiveness.} The reflective feedback includes different reasoning outputs when multiple agents are queried, which can help formalise a comprehensive refinement suggestion.

  \item \textit{Scalability.} \textit{Cross-reflection} supports scalable agent-based systems as the reflective agents can be flexibly updated without disrupting the system operation.

%  \item \textit{Continuous improvement.} The agent can continuously update the knowledge base and the manner of formalising the prompts and knowledge, to provide more reliable and coherent output to users without or with fewer reflection steps.

\end{itemize}

Drawbacks: 
\begin{itemize}
   \item \textit{Reasoning uncertainty.} The overall response quality and reliability are dependent on the performance of other reflective agents.

%   \item \textit{Security protection.} The generated response may be leaked during transmission or the reflective agents.

   \item \textit{Fairness preservation.} When various agents participate in the reflection process, a critical issue would be how to preserve fairness among all the provided feedback.

   \item \textit{Complex accountability.} If the cross-reflection feedback causes serious or harmful results, the accountability process may be complex when multiple agents are employed.

   \item \textit{Overhead.} i) There will be communication overhead for the interactions between agents. ii) Users may need to pay for utilising the reflective agents.
\end{itemize}

\vspace{0.5em}\noindent \textbf{Known uses:} 
\begin{itemize}

   \item \textit{XAgent}\footnote{\url{https://github.com/OpenBMB/XAgent}\label{XAgent}}. In XAgent, the tool agent can send feedback and reflection to the plan agent to indicate whether a task is completed, or pinpoint the refinements.
   
   \item Yao et al.~\cite{yao2023retroformer} explore agents' capability of learning through communicating with each other. A thinker agent can provide suggestions to an actor agent, who is responsible for decision-making.

   \item Qian et al.~\cite{qian2023communicative} develop a virtual software development company based on agents, where the tester agents can detect bugs and report to programmer agents.

   \item Talebirad and Nadiri~\cite{talebirad2023multi} analyse the inter-agent feedback which involves criticism of each other, which can help agents adapt their strategies.

   %https://arxiv.org/abs/2404.18796
\end{itemize}

\vspace{0.5em}\noindent \textbf{Related patterns:} 

\begin{itemize}
    \item \textit{Prompt/response optimiser.} \textit{Cross-reflection} can provide feedback to improve the output of \textit{prompt/response optimiser}.

    \item \textit{Voting-based, role-based,} and \textit{debate-based cooperation.} Reflective agents can collaborate to evaluate an agent's outputs in different cooperation schemes.
    
    \item \textit{Tool/agent registry.} The agent can search reflective agents for \textit{cross-reflection} via \textit{tool/agent registry}.
\end{itemize}

\subsection{Human Reflection}

\vspace{0.5em}\noindent \textbf{Summary:} The agent collects feedback from humans to refine the plan, to effectively align with the human preference.

\vspace{0.5em}\noindent \textbf{Context:} Agents create plans and strategies that decompose users' goals and requirements into a pool of tasks. The tasks will be completed by other tools and agents.

% In the user feedback evaluation, once the agent has indicated completion of the task from the original input dialogue, the agent will query feedback from the user. 

% If the simulator indicates success of the task, the agent will end the episode. If the simulator indicates the task is not successful, feedback will be given to the agent for additional planning.

\vspace{0.5em}\noindent \textbf{Problem:} How to ensure human preference is fully and correctly captured and integrated into the reasoning process and generated plans?
%If human instructions accumulate over longer multi-step interactions that fall outside the receding context horizon, previous instructions can simply be forgotten.

\vspace{0.5em}\noindent \textbf{Forces:} 

\begin{itemize}
  \item \textit{Alignment to human preference.} Agents are expected to achieve users' goals ultimately, consequently, it is critical for agents to comprehend users' preferences.

  \item \textit{Contestability.} If the agent's outputs do not satisfy users' requirements and will cause negative impacts, there should be a timely process for users to contest the responses of agent.
\end{itemize}

\vspace{0.5em}\noindent \textbf{Solution:} Fig.~\ref{fig:reflection} presents a high-level graphical representation of \textit{human-reflection}. When a user prompts his/her goals and specified constraints, the agent first creates a plan consisting of a series of intermediate steps. The constructed plan and its reasoning process logs can be presented to the user for review, or sent to other human experts to validate the feasibility and usefulness. The user or expert can provide comments or suggestions to indicate which steps can be updated or replaced. The plan will be iteratively assessed and improved until it is approved by the user/expert.

\vspace{0.5em}\noindent \textbf{Consequences:} 

Benefits:
\begin{itemize}
  % \item \textit{Performance.} Asking for a user’s feedback twice improves performance by 1.27x. 
  % For example, LMPC can learn to learn faster from human feedback, and we observe performance gains on test tasks and test robot embodiments.\liming{Learning to Learn Faster from Human Feedback with Language Model Predictive Control}
  % The paper seems more of multi-plan generator

  \item \textit{Alignment to human preference.} The agent can directly receive feedback from users or additional human experts to understand human preferences, and improve the outcomes or procedural fairness, diversity in the results, etc.

  \item \textit{Contestability.} Users or human experts can challenge the agent's outcomes immediately if abnormal behaviours or responses are found.

  \item \textit{Effectiveness.} \textit{Human-reflection} allows agents to include users' perspectives for plan refinement, which can help formalise responses tailored to users' specific needs and level of understanding. This can ensure the usability of strategies, and improve the effectiveness for achieving users' goals.

%  \item \textit{Fairness.} Agents can absorb and consider users' background, experiences, and potential biases via \textit{human-reflection}, to assess and mitigate the cognitive biases of their own.

%  \item \textit{Continuous improvement.} The agent can continuously update the knowledge base and the manner of formalising the prompts and knowledge, to provide more reliable and coherent output to users without or with fewer reflection steps.
\end{itemize}

Drawbacks: 
\begin{itemize}
   \item \textit{Fairness preservation.} The agent may be affected by users who provide skewed information about the real world. 

   \item \textit{Limited capability.} Agents may still have limited capability to understand human emotions and experiences.

   \item \textit{Underspecification.} Users may provide limited or ambiguous reflective feedback to agents.

   \item \textit{Overhead.} Users may need to pay for the multiple rounds of communication with the agent.
\end{itemize}

\vspace{0.5em}\noindent \textbf{Known uses:} 
\begin{itemize}
   \item \textit{Inner Monologue}~\cite{huang2022inner}. Inner Monologue is implemented in a robotic system, which can decompose users' instructions into actionable steps, and leverage human feedback for object recognition.

   \item Ma et al.~\cite{ma2024towards} explore the deliberation between users and agents for decision-making. Users and agents both need to provide related evidence and arguments for their conflicting opinions.

   \item Wang et al.~\cite{wang2023drdt} incorporate human feedback for agents to capture the dynamic evolution of user interests and consequently provide more accurate recommendations.
\end{itemize}

\vspace{0.5em}\noindent \textbf{Related patterns:} 

\begin{itemize}
    \item \textit{Prompt/response optimiser.} \textit{Human-reflection} can provide human preference and suggestions to improve the generated prompts and responses.

    \item \textit{Multi-path plan generator.} \textit{Multi-plan generator} creates plans with various directions, and \textit{human-reflection} can help finalise the plan with user feedback to determine the choice of each intermediate step.

    \item \textit{Incremental model querying.} \textit{Human-reflection} is enabled by \textit{incremental model querying} for iterative communication between users/experts and the agent.
\end{itemize}

\subsection{Voting-based Cooperation}

\vspace{0.5em}\noindent \textbf{Summary:} Agents can freely provide their opinions and reach consensus through voting-based cooperation.

\vspace{0.5em}\noindent \textbf{Context:} Multiple agents can be leveraged within a compound AI system. Agents need to collaborate on the same task while having their own perspectives.

\vspace{0.5em}\noindent \textbf{Problem:} How to finalise the agents' decisions properly to ensure fairness among different agents?

\vspace{0.5em}\noindent \textbf{Forces:} 

\begin{itemize}
  \item \textit{Diversity.} The employed agents can have diverse opinions of how a plan is constructed or how a task should be completed.

  \item \textit{Fairness.} Decision-making among agents should take their rights and responsibilities into consideration to preserve fairness.

  \item \textit{Accountability.} The behaviours of agents should be recorded to enable future auditing if any violation is found in the collaboration outcomes.
\end{itemize}

\begin{figure}[!ht]
    \centering
    \includegraphics[width=0.7\columnwidth]{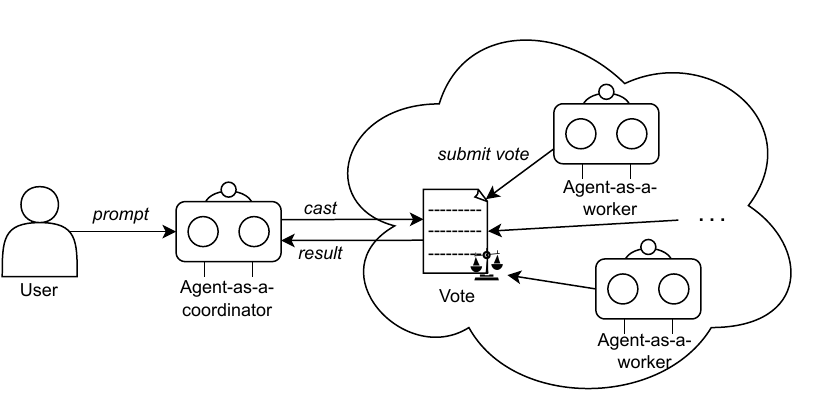}	
    \caption{Voting-based cooperation.}
    \label{fig:voting}
\end{figure}

\vspace{0.5em}\noindent \textbf{Solution:} Fig.~\ref{fig:voting} illustrates how agents can cooperate to finalise a decision via votes. Specifically, an agent can first generate a candidate response to the user's prompts, then it holds a vote in which different reflective suggestions are presented as choices. Additional agents are requested to submit their votes to select the most appropriate feedback according to their capabilities and experiences. In this circumstance, agents communicate in a centralised manner that the original agent will act as a coordinator. The voting result will be formalised and sent back to the original agent, who can refine the response accordingly before answering the user. Please note that the voting process can be implemented through various methods, e.g., direct communications between the agent-as-a-coordinator and other agents, blockchain-based smart contracts, etc. Moreover, the identity management of all participating agents is significant to ensure the traceability and verifiability of votes.

%The agent queries a coordinator agent for reflective feedback, while the latter agent holds a vote in which different reflective suggestions are presented as choices. Additional agents are requested to submit their votes to select the most appropriate feedback according to their capabilities and experiences. The voting result will be formalised and sent back to the original agent, who can refine the response accordingly before answering the user.

\vspace{0.5em}\noindent \textbf{Consequences:} 

Benefits:
\begin{itemize}
%  \item \textit{Decision finalisation.} Agents can reach consensus on responses or task results from diverse perspectives.

  \item \textit{Fairness.} Votes can be held in multiple ways to preserve fairness. For instance, counting heads to ensure agents' rights are equal, or weights can be distributed considering the roles of agents, etc.

  \item \textit{Accountability.} The overall procedure and final results are recorded in the respective voting system. Stakeholders can trace back to identify the accountable agents selecting certain options.

  \item \textit{Collective intelligence.} The finalised decisions after votes can leverage the strengths of multiple agents (e.g. comprehensive knowledge base), hence they are regarded as more accurate and reliable than the ones generated by a single agent.

  %\item \textit{Availability.} When multiple agents are leveraged, the system can continue operation in case that some agents become inactive. 

\end{itemize}

Drawbacks: 
\begin{itemize}
   \item \textit{Centralisation.} Specific agents may gain the majority of decision rights and hence have the ability to compromise the voting process.

   \item \textit{Overhead.} Hosting a vote may increase the communication overhead for agents to examine and vote for the choices. 
\end{itemize}

\vspace{0.5em}\noindent \textbf{Known uses:} 
\begin{itemize}
   \item Hamilton~\cite{hamilton2023blind} utilises nine agents to simulate court where the agents need to vote for the received cases. Each case is determined by the dominant voting result.

   \item ChatEval~\cite{chan2024chateval}. Agents can reach consensus on users' prompts via voting, while the voting results can be totalled by calculating either the majority vote or the average score.

   \item Yang et al.~\cite{yang2024llm} explore the alignment of agent voters based on GPT-4 and LLaMA-2 and human voters on 24 urban projects. The results indicate that agent voters tend to have uniform choices while human voters have diverse preferences.

   %https://arxiv.org/pdf/2402.05120

   \item Li et al.~\cite{li2024more} incrementally query a foundation model to generate $N$ samples, and leverage multiple agents to select a finale response via majority voting. 

\end{itemize}

\vspace{0.5em}\noindent \textbf{Related patterns:} 

\begin{itemize}
    \item \textit{Cross-reflection.} An agent can query multiple agents to provide feedback, which can be determined via \textit{voting-based cooperation} between the reflective agents.

    \item \textit{Role-based} and \textit{debate-based cooperation}. \textit{Voting-based cooperation} can be regarded as an alternative to other cooperation patterns by hosting a vote between agents, whilst they can be applied together to complement each other.

    \item \textit{Tool/agent registry}. Agents participating in the voting process can be employed via \textit{tool/agent registry}.
\end{itemize}

\subsection{Role-based Cooperation}

\vspace{0.5em}\noindent \textbf{Summary:} Agents are assigned assorted roles and decisions are finalised in accordance with their roles.

\vspace{0.5em}\noindent \textbf{Context:} Multiple agents can be leveraged within a compound AI system. Agents need to collaborate on the same task while having their own perspectives.

%The roles are played by different agents. This pattern also supports communicating with external tools and other agents, which can be easily achieved by changing the roles’ prompts or replacing the agents behind the roles.

\vspace{0.5em}\noindent \textbf{Problem:} How can agents cooperate on certain tasks considering their specialties?

\vspace{0.5em}\noindent \textbf{Forces:} 

\begin{itemize}
  \item \textit{Diversity.} The employed agents can have diverse opinions of how a plan is constructed or how a task should be completed.

  \item \textit{Division of labor.} As agents can be trained with different corpus for various purposes, their strengths and expertise should be taken into consideration for task completion.

  \item \textit{Fault tolerance.} Agents may be unavailable during cooperation, which will affect the eventual task result.
\end{itemize}

\begin{figure}[!ht]
    \centering
    \includegraphics[width=0.7\columnwidth]{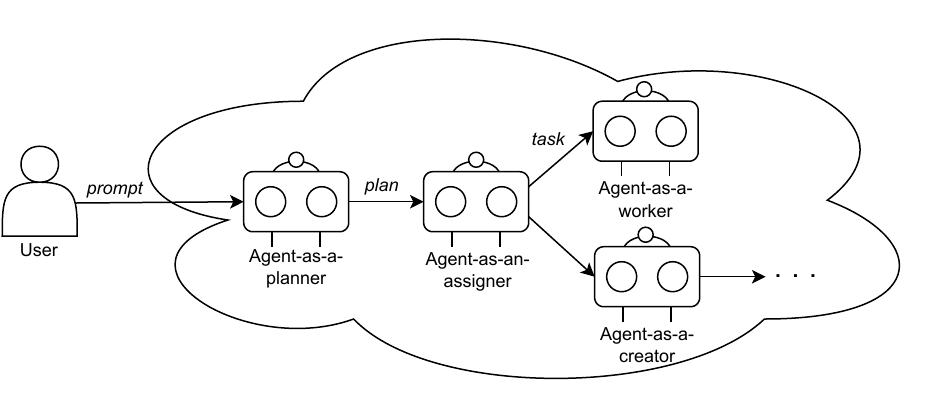}	
    \caption{Role-based cooperation.}
    \label{fig:role}
\end{figure}

\vspace{0.5em}\noindent \textbf{Solution:} Fig.~\ref{fig:role} illustrates a high-level graphical representation of \textit{role-based cooperation}, where agents coordinate in a hierarchical scheme. In particular, agents can be assigned certain roles and hence establishing a workflow via persona specification, task definition, tool employment, and process orchestration. For example, an agent-as-a-planner can generate a multi-step plan by decomposing user's goal into a chain of tasks. Subsequently, the agent-as-an-assigner can orchestrate task assignment, i.e., some tasks can be completed by the assigner itself, while other tasks can be delegated to certain agent-as-a-worker based on their domain-specific capabilities and expertise. In addition, if there is no available agent, agent-as-a-creator can be invoked to create a new agent with a specific role, by providing necessary resources, clear objectives and initial guidance to ensure a seamless transition of tasks and responsibilities. Please note that more elaborate roles can be defined and assigned to the agents.

%In particular, an agent creator can assign a specific role to a newly created agent with the necessary resources to enable specific capabilities and expertise, by providing clear objectives and initial guidance to ensure a seamless transition of tasks and responsibilities. When multiple agents collaboratively work on certain tasks, they should all operate autonomously within the bounds of their assigned roles. For instance, the chef agent, barista agent, and waiter agent can run a virtual café together, while agents with new roles can be introduced anytime to extend the capability of the whole system.

%When a new agent is created, the creator assigns the new agent a role, grants it the necessary properties, and establishes connections with other agents and plugins. These properties and connections are subsets of those available to the creator agent. Also, a connection to the creator is established. Once the new agent is created and initialized, it operates independently within its defined role. The creator agent sets a clear goal for the new agent, providing initial guidance to ensure a smooth transition of responsibilities. By allowing agents to dynamically create new agents and delegate tasks, the system can effectively manage workloads, enhance parallel processing capabilities, and improve overall system performance. This dynamic approach fosters a collaborative environment where agents can dynamically organize and distribute tasks, ultimately contributing to the achievement of the common goal.

\vspace{0.5em}\noindent \textbf{Consequences:} 

Benefits:
\begin{itemize}
%    \item \textit{Decision finalisation.} Agents can complete the tasks and generate responses collaboratively based on their roles.

    \item \textit{Division of labor.} Agents can simulate the division of labor in the real world according to their roles, which enables the observation of social phenomena.

    \item \textit{Fault tolerance.} Since multiple agents are leveraged, the system can continue operation by replacing inactive agents with other agents of the same role. 

    \item \textit{Scalability.} Agents of new roles can be employed or created anytime to refine the task workflow and extend the capability of the whole system.
  %[2]delegating different roles to each agent introduces more flexibility and efficiency in the context task management. the agents’ ability to dynamically adapt and distribute workload ensures flexibility and resilience under changing conditions and demands.

    \item \textit{Accountability.} Accountability is facilitated as the responsibilities of agents are attributed clearly regarding their expected roles.
\end{itemize}

Drawbacks: 
\begin{itemize}
    \item \textit{Overhead.} Cooperation between agents will increase communication overhead, while agent services with different roles may have different prices.
\end{itemize}

\vspace{0.5em}\noindent \textbf{Known uses:} 
\begin{itemize}
   % \item \textit{[1]autonomous software development multiagent framework.} 
   % Qian et al. [31] and Hong et al. [14] present multi-agent frameworks for autonomous software development, where agents play different roles in a predetermined workflow, such as programmer and project manager
   
   % \item \textit{AutoGPT.} 
   % As another example, one can implement a concept of co-agents, where multiple autonomous instances of Auto-GPT could collaborate, share a workspace for files, and communicate in a board, essentially mimicking a team of humans working remotely, with each having a specific role.

   % MedAgents: Large Language Models as Collaborators for Zero-shot Medical Reasoning

   % Retroformer: retrospective large language agents with policy gradient optimization

   % Communicative Agents for Software Development

   % Selective Reflection-Tuning: Student-Selected Data Recycling for LLM Instruction-Tuning

   % \liming{1.MetaAgents: Simulating Interactions of Human Behaviors for LLM-based Task-oriented Coordination via Collaborative Generative Agents;2.MULTI-AGENT COLLABORATION: HARNESSING THE POWER OF INTELLIGENT LLM AGENTS}

   % GPTs

   \item \textit{XAgent}\textsuperscript{\ref{XAgent}}. XAgent consists of three main parts: planner agent for task generation, dispatcher agent for task assignment, and tool agent for task completion.
   
   \item \textit{MetaGPT}~\cite{hong2023metagpt}. MetaGPT utilises various agents acting as different roles (e.g., architect, project manager, engineer) in standardized operating procedures.

   \item \textit{MedAgents}~\cite{tang2023medagents}. Agents are assigned roles as various domain experts (e.g. cardiology, surgery, gastroenterology) to provide specialised analysis and collaboratively work on healthcare issues.

   \item Wang et al.~\cite{wang2024mixture} propose Mixture-of-Agents where proposer agents provide useful reference responses to aggregator agents, and the aggregator agents are composed in layers to synthesise and refine the responses.

   %\item Li et al.~\cite{li2024selective} propose selective reflection-tuning in which the teacher agent's reflection on data quality and student agent's data selection are synergised.

   %https://arxiv.org/abs/2402.14207
   
\end{itemize}

\vspace{0.5em}\noindent \textbf{Related patterns:} 

\begin{itemize}
    \item \textit{Cross-reflection.} An agent can query multiple agents to provide feedback, which can be determined via \textit{role-based cooperation} between the reflective agents.

    \item \textit{Voting-based} and \textit{debate-based cooperation}. \textit{Role-based cooperation} can be regarded as an alternative of other cooperation patterns by clearly assigning roles to agents, which will then work and collaborate according to the given roles. Whilst, these patterns can be applied together to complement each other.

    \item \textit{Tool/agent registry}. Agents with different roles can be searched and employed via \textit{tool/agent registry}.
\end{itemize}

\subsection{Debate-based Cooperation}

\vspace{0.5em}\noindent \textbf{Summary:} An agent receives feedback from other agents, and adjusts the thoughts and behaviours during the debate with other agents until a consensus is reached.

%\liming{1.Improving Factuality and Reasoning in Language Models through Multiagent Debate;2.Encouraging Divergent Thinking in Large Language Models through Multi-Agent Debate;3.Improving Factuality and Reasoning in Language Models}

\vspace{0.5em}\noindent \textbf{Context:} A compound AI system can integrate multiple agents to provide more comprehensive services. The included agents need to collaborate on the same task while having their own perspectives.

% An extensive body of recent work has focused on improving factual accuracy and reasoning in language models. These range from prompting models with few or zero-shot chain-of-thought demonstrations, use of verification, self-consistency, or intermediate scratchpads.
% We note that these techniques are applied over a single model instance. Instead, we propose a complementary approach inspired by The Society of Mind [19] and multi-agent settings, where multiple language model instances (or agents) individually propose and jointly debate their responses and reasoning processes to arrive at a single common answer.

\vspace{0.5em}\noindent \textbf{Problem:} How to leverage multiple agents to create refined responses, while facilitating the evolution of agents.

%[3]While the debate process is more costly, requiring multiple model instances and rounds, it arrives at significantly improved answers and may be used to generate additional model training data, effectively creating a model self-improvement loop.

\vspace{0.5em}\noindent \textbf{Forces:} 

\begin{itemize}
     \item \textit{Diversity.} Different agents may have various opinions to help refine the generated responses to users.

     \item \textit{Lack of adaptability.} An agent may exhibit limited creativity in reasoning and response generation when given new context or tasks.

     \item \textit{Lack of explainability.} The interaction process of agents should be interpreted for auditing if violations are detected.
\end{itemize}

\begin{figure}[!ht]
    \centering
    \includegraphics[width=0.6\columnwidth]{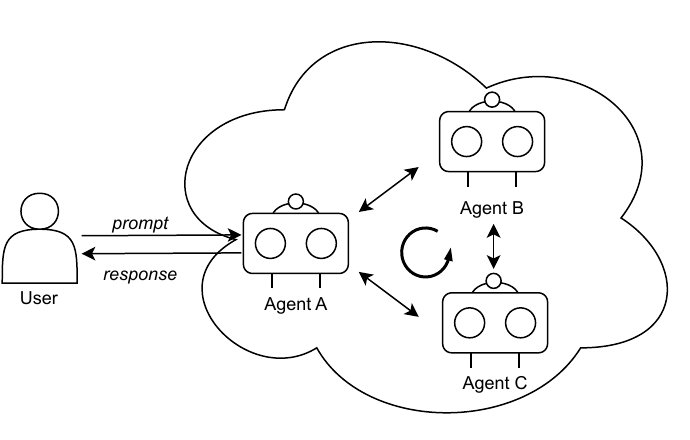}	
    \caption{Debate-based cooperation.}
    \label{fig:debate}
\end{figure}

\vspace{0.5em}\noindent \textbf{Solution:} Fig.~\ref{fig:debate} depicts a high-level graphical representation of \textit{debate-based cooperation}. A user can send queries to an agent, which will then share the questions with other agents. Given the shared question, each agent generates its own initial responses, and subsequently, a round of debate will start between the agents. Agents will propagate their initial response in a decentralised manner to each other for verification, while also providing instructions and potential directions to construct a more comprehensive response based on inclusive and collective outcomes. In addition, agents may utilise a shared memory in certain circumstances, or allow each other to access the respective memory for facilitating the debate. This debate process can be iterative to enhance the performance of all participating agents. \textit{Debate-based cooperation} can end according to a predefined number of debate rounds, or the agents will continue the procedure until a consensus answer is obtained.

\vspace{0.5em}\noindent \textbf{Consequences:} 

Benefits:
\begin{itemize}

    \item \textit{Adaptability.} Agents can adapt to other agents during the debate procedure, achieving continuous learning and evolution.

    \item \textit{Explainability.} \textit{Debate-based cooperation} is structured with agents' arguments and presented evidence, preserving transparency and explainability of the whole procedure.

    %\item \textit{Response accuracy.} During the cooperation, agents can provide accurate and up-to-date knowledge to each other, ensuring the reliability of eventual responses.

    \item \textit{Critical thinking.} Arguing with other agents can help an agent develop the ability of critical thinking for future reasoning process.

\end{itemize}

Drawbacks: 
\begin{itemize}
   \item \textit{Limited capability.} The effectiveness of \textit{debate-based cooperation} relies on agents' capabilities of reasoning, argument, and evaluation of other agents' statement.

   \item \textit{Data privacy.} Agents may need to withhold certain sensitive information, which can affect the debate process.
   
   \item \textit{Overhead.} The complexity of debate may increase the communication and computation overhead.

   \item \textit{Scalability preservation.} The system scalability may be affected as the number of participating agents increases. The coordination of agents and processing of their arguments may become complex.
\end{itemize}

\vspace{0.5em}\noindent \textbf{Known uses:} 
\begin{itemize}
   \item \textit{crewAI}\footnote{\url{https://www.crewai.com/}}. crewAI provides a multi-agent orchestration framework where multiple agents can be grouped for discussion on a given topic.
   
   \item Liang et al.~\cite{liang2023encouraging} leverage multi-agent debate to address the issue of ``Degeneration-of-Thought''. Within the debate, an agent needs to persuade another and correct the mistakes.

   \item Du et al.~\cite{du2023improving} employ multiple agents to discuss the given user input, and the experiment results indicate that the agents can converge on a consensus answer after multiple rounds.

   \item Chen et al.~\cite{chen2023multi} explore the negotiation process in a multi-agent system, where each agent can perceive the outcomes of other agents, and adjust its own strategies.

   \item Li et al.~\cite{li2023prd} propose a framework including peer rank and discussion between agents, to mitigate the biases in automated evaluation process.

   %https://arxiv.org/pdf/2311.11855
\end{itemize}

\vspace{0.5em}\noindent \textbf{Related patterns:} 

\begin{itemize}
    \item \textit{Cross-reflection.} Agents can decide the reflective feedback to another agent via \textit{debate-based cooperation}.

    \item \textit{Voting-based} and \textit{role-based cooperation}. \textit{Debate-based cooperation} can be regarded as an alternative of other cooperation patterns by hosting a debate between agents, whilst they can be applied together to complement each other.

    \item \textit{Tool/agent registry}. Agents participating in the debate process can be employed via \textit{tool/agent registry}.
\end{itemize}

\subsection{Multimodal Guardrails}

\vspace{0.5em}\noindent \textbf{Summary:} Multimodal guardrails can control the inputs and outputs of foundation models to meet specific requirements such as user requirements, ethical standards, and laws.

\vspace{0.5em}\noindent \textbf{Context:} An agent consists of foundation model and other components. When users prompt specific goals to the agent, the underlying foundation model is queried for goal achievement.

\vspace{0.5em}\noindent \textbf{Problem:} How to prevent the foundation model from being influenced by adversarial inputs, or generate harmful or undesirable outputs to users and other components?

\vspace{0.5em}\noindent \textbf{Forces:} 

\begin{itemize}
  \item \textit{Robustness.} Adversarial information may be sent to the foundation model, which will affect the model's memory and all subsequent reasoning processes and results.

  \item \textit{Safety.} Foundation models may generate inappropriate responses due to hallucinations, which can be offensive to users, and disturb the operation of other components (e.g., other agents, external tools).

  \item \textit{Standard alignment.} Agents and the underlying foundation models should align with the specific standards and requirements in industries and organisations.
\end{itemize}

\begin{figure}[t]
    \centering
    \includegraphics[width=0.8\columnwidth]{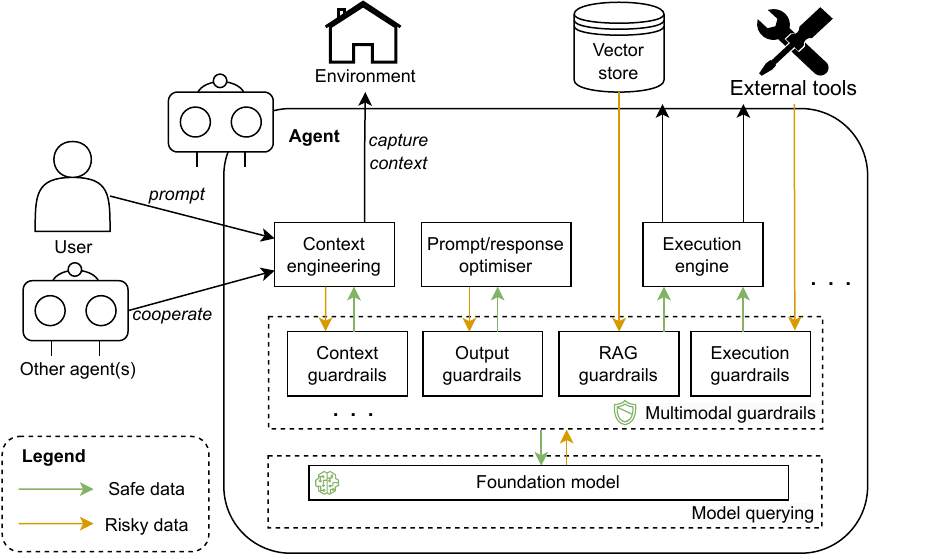}	
    \caption{Multimodal guardrails.}
    \label{fig:guardrail}
\end{figure}

\vspace{0.5em}\noindent \textbf{Solution:} Fig.~\ref{fig:guardrail} presents a simplified graphical representation of \textit{multimodal guardrails}. Guardrails can be applied as an intermediate layer between the foundation model and all other components in a compound AI system. When users input prompts or other components (e.g. memory) send any message to the foundation model, guardrails can first verify whether the information meets specific predefined requirements. Only valid information are delivered to the foundation model, while risky or sensitive data will be processed before being transferred. For instance, personally identifiable information should be treated with care or removed to protect privacy. Guardrails can evaluate the contents either relying on predefined examples, or in a ``reference-free'' manner. Equivalently, when the foundation model creates results, guardrails need to ensure that the responses do not include biased or irrespective information to users, or fulfil the particular requirements of other system components. Please note that a set of guardrails can be implemented where each of them is responsible for specialised interactions, e.g., information retrieval from datastore, validation of users' input, external API invocation, etc. Meanwhile, guardrails are capable of processing multimodal data such as text, audio, video to provide comprehensive monitoring and control.

\vspace{0.5em}\noindent \textbf{Consequences:} 

Benefits:
\begin{itemize}

  \item \textit{Robustness.} Guardrails preserve the robustness of foundation models by filtering the inappropriate context information.

  \item \textit{Safety.} Guardrails serve as validators of foundation model outcomes, ensuring the generated responses do not harm agent users.

  \item \textit{Standard alignment.} Guardrails can be configured referring to organisational policies and strategies, ethical standards, and legal requirements to regulate the behaviours of foundation models.

  \item \textit{Adaptability.} Guardrails can be implemented across various foundation models and agents, and deployed with customised requirements.
\end{itemize}

Drawbacks: 
\begin{itemize}
    \item \textit{Overhead.} i) Collecting diverse and high-quality corpus to develop \textit{multimodal guardrails} may be resource-intensive. ii) Real-time processing multimodal data can increase the computational requirements and costs.

    \item \textit{Lack of explainability.} The complexity of \textit{multimodal guardrails} makes it difficult to explain the finalised outputs.

\end{itemize}

\vspace{0.5em}\noindent \textbf{Known uses:} 
\begin{itemize}

   \item \textit{NeMo guardrails}~\cite{rebedea2023nemo}. NVIDIA released NeMo guardrails, which are specifically designed to ensure the coherency of dialogue between users and AI systems, and prevent negative impact of misinformation and sensitive topics.

   \item \textit{Llama guard}~\cite{inan2023llama}. Meta published Llama guard, a foundation model based safeguard model fine-tuned via a safety risk taxonomy. Llama guard can identify the potentially risky or violating content in users' prompts and model outputs.

   \item \textit{Guardrails AI\footnote{\url{https://www.guardrailsai.com/}}.} Guardrails AI provides a hub, listing various validators for handling different risks in the inputs and outputs of foundation models.
   
\end{itemize}

\vspace{0.5em}\noindent \textbf{Related patterns:} 

\begin{itemize}
    \item \textit{Proactive goal creator.} \textit{Multimodal guardrails} can help process the multimodal data captured by \textit{proactive goal creator}.
    
    \item \textit{One-shot} and \textit{incremental model querying}. \textit{Multimodal guardrails} serve as an intermediate layer, managing the inputs and outputs of model querying.
    
\end{itemize}

\subsection{Tool/Agent Registry}

\vspace{0.5em}\noindent \textbf{Summary:} The tool/agent registry maintains a unified and convenient source to select diverse agents and tools.

\vspace{0.5em}\noindent \textbf{Context:} Within an agent, the task executor may cooperate with other agents or leverage external tools for expanded capabilities.

\vspace{0.5em}\noindent \textbf{Problem:} There are diverse agents and tools, how can the agent efficiently select the appropriate external agents and tools?

\vspace{0.5em}\noindent \textbf{Forces:} 

\begin{itemize}  
  \item \textit{Discoverability.} It may be difficult for users and agents to discover the available agents and tools considering the diversity.

  \item \textit{Efficiency.} Users/agents need to finalise agent and tool selection within a certain time period.

  \item \textit{Tool appropriateness.} Particular tasks may have specific requirements of agents/tools (e.g. certain capabilities).
\end{itemize}

\begin{figure}[!ht]
    \centering
    \includegraphics[width=0.75\columnwidth]{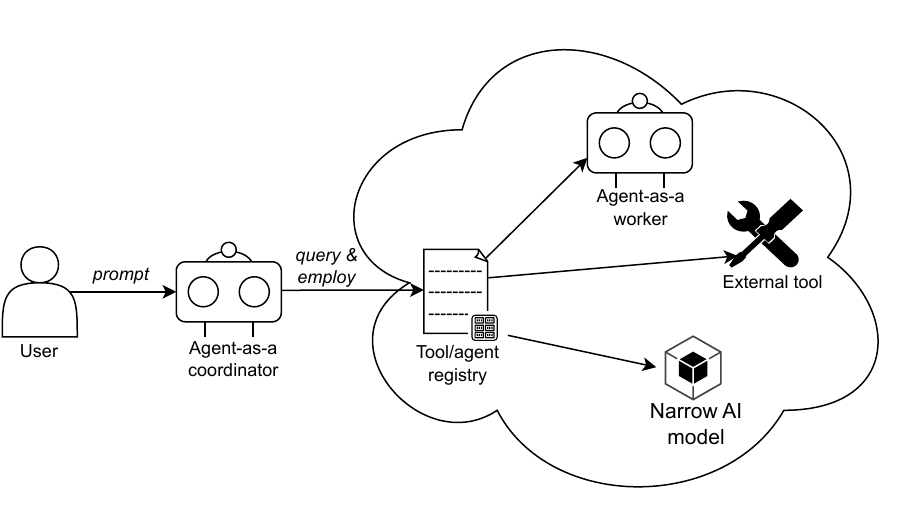}	
    \caption{Tool/agent registry.}
    \label{fig:registry}
\end{figure}

\vspace{0.5em}\noindent \textbf{Solution:} Fig.~\ref{fig:registry} depicts how an agent searches external agents and tools via a \textit{tool/agent registry}. A user prompts goals to an agent, which then decomposes the goals into fine-grained tasks. The agent can query the \textit{tool/agent registry}, which is the main entry point for collecting and categorising various tools and agents regarding a series of metrics (e.g., domain-specific capability, price, context window). Based on the returned information, the agent can employ and assign the tasks to respective tools and agents. Please note that a registry can be implemented in different manners, for instance, a coordinator agent with specific knowledge base, blockchain-based smart contract, etc., and a registry can be extended into a marketplace for tool/agent service trading.

\vspace{0.5em}\noindent \textbf{Consequences:} 

Benefits:
\begin{itemize}
  \item \textit{Discoverability.} The registry provides a catalogue for users and agents to discover tools and agents with different capabilities.

  \item \textit{Efficiency.} The registry offers an intuitive inventory listing the attributes (e.g., performance, price) of tools and agents, which saves time for comparison.

  \item \textit{Tool appropriateness.} Given the task requirements and conditions, users and agents can select the most appropriate tools/agents according to the provided attributes.

  \item \textit{Scalability.} The registry only stores certain metadata about tools and agents, hence the data structure is simple and lightweight, which ensures the scalability of the registry.
\end{itemize}

Drawbacks: 
\begin{itemize}
   \item \textit{Centralisation.} The registry may become a vendor lock-in solution and cause single point of failure. It may be manipulated and compromised if it is maintained by external entities.

   \item \textit{Overhead.} Implementing and maintaining a tool/agent registry can introduce additional complexity and overhead.
\end{itemize}

\vspace{0.5em}\noindent \textbf{Known uses:} 
\begin{itemize}
   \item \textit{GPTStore}\footnote{\url{https://gptstore.ai/}}. GPTStore provides a catalogue for searching ChatGPT-based agents.
   
   \item \textit{TPTU}~\cite{ruan2023tptu}. TPTU incorporates a toolset to broaden the capabilities of AI Agents.

   \item \textit{VOYAGER}~\cite{wang2023voyager}. VOYAGER can store action programs and hence incrementally establish a skill library for reusability.

   \item \textit{OpenAgents}~\cite{xie2023openagents}. An agent is specifically developed to manage the API invocation of plugins. 

   %https://arxiv.org/pdf/2306.06624
   %https://arxiv.org/abs/2305.15334
   %https://arxiv.org/abs/2307.16789
\end{itemize}

\vspace{0.5em}\noindent \textbf{Related patterns:} 

\begin{itemize}
    \item \textit{Cross-reflection.} The agent can search reflective agents for \textit{cross-reflection} via \textit{tool/agent registry}.

    \item \textit{Voting-based, role-based} and \textit{debate-based cooperation.} \textit{Tool/agent registry} can provide a source of agents for the cooperation patterns.

    \item \textit{Agent adapter.} \textit{Tool/agent registry} records the available external tools, while \textit{agent adapter} can convert the interface of selected tools into agent-friendly format.
\end{itemize}

\subsection{Agent Adapter}
\vspace{0.5em}\noindent \textbf{Summary:} An agent adapter provides interface to connect the agent and external tools for task completion.

%An agent adapter can connect the agent and external tools by learning new interfaces and converting incompatible interfaces into expected ones.

\vspace{0.5em}\noindent \textbf{Context:} An agent may leverage external tools to complete certain tasks for expanded capabilities.

\vspace{0.5em}\noindent \textbf{Problem:} The agent needs to deal with different interfaces of diverse tools, while certain interfaces might be incompatible or inefficient to interact for the agent. How can the agent assign tasks to external tools and process the results?

%The agent/foundation model and tool interfaces may be incompatible, how can the agent assign tasks to external tools and process the results?

\vspace{0.5em}\noindent \textbf{Forces:} 
\begin{itemize}
    \item \textit{Interoperability.} Certain tasks require external tools to complete, and the tools may need agents to process particular information during intermediate steps.

    \item \textit{Adaptability.} Agents may employ new tools considering task complexity, tool capability, cost, etc.

    \item \textit{Overhead.} Manual development of compatible interfaces for agents and external tools can be intensive and inefficient.

\end{itemize}

\begin{figure}[!ht]
    \centering
    \includegraphics[width=0.56\columnwidth]{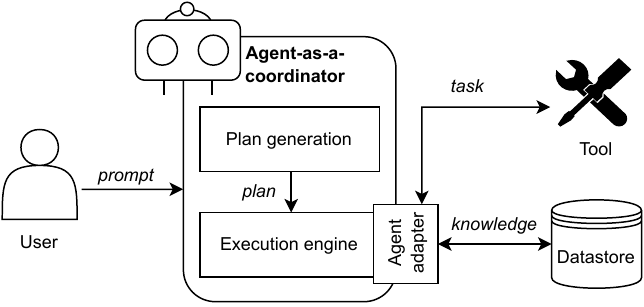}	
    \caption{Agent adapter.}
    \label{fig:adapter}
\end{figure}

\vspace{0.5em}\noindent \textbf{Solution:} Fig.~\ref{fig:adapter} demonstrates a simplified graphical representation of \textit{agent adapter}. Given user's instructions, the agent generates a plan consisting of a set of tasks to achieve the user's goals. In particular, the agent may employ diverse external tools to complete different tasks. However, tools have respective interfaces, which can be of different abstraction levels for the agent to deal with, or have specific format requirements, etc. \textit{Agent adapter} can help invoke and manage these interfaces by converting the agent messages into required format or content, and vice versa. In particular, the adapter can retrieve tool manual or tutorial from datastore, to acquire available interfaces and learn the usage. It then transforms the agent outputs based on the interface requirements and invokes the service~\cite{qu2024tool}. Please note that fine-grained interface description can enhance agent understanding and hence improve the performance. The adapter also receives execution results from tools, which will be sent to the underlying foundation model for further analysis (e.g. task assignment to other tools, self-reflection for tool employment). For instance, the adapter can translate tasks into system messages when interacting with local file system, or capture and operate graphical user interface when playing a video game.

%sending system messages when interacting with the file system, capturing graphical user interface when playing a video game. In addition, it can retrieve specific manual or tutorial from datastore to ensure the correctness of leveraging tool interface. 

%needs to acquire available interface from 

%the agent and underlying foundation model learn how to invoke and manage these interfaces, or convert the interface in a way that the agent can interact with, for instance, sending system messages when interacting with the file system, capturing graphical user interface when playing a video game. In addition, it can retrieve specific manual or tutorial from datastore to ensure the correctness of leveraging tool interface. 

 %functioning as an interface adapter for software components that use different data formats to communicate with each other, e.g. parse the task into machine-readable template for executing by an AI model.

% high/low level
% operating system
% graphical ui
% retrieve specific manual tutorial

\vspace{0.5em}\noindent \textbf{Consequences:} 

Benefits:
\begin{itemize}
    \item \textit{Interoperability.} \textit{Agent adapter} facilitates the interoperation between an agent and external tools.

    \item \textit{Adaptability.} Agents can employ new tools via \textit{agent adapter}, which can acquire and convert the tool API via corresponding manual or tutorial.
    
    %learn the manual of tool interfaces to ensure compatibility.

    \item \textit{Reduced development cost.} \textit{Agent adapter} enables autonomous conversion of interfaces, there is no need to develop compatible interfaces for different tools, hence the development cost is reduced.

\end{itemize}

Drawbacks:

\begin{itemize}
    \item \textit{Maintenance overhead.} i) \textit{Agent adapter} itself requires proper maintenance and evaluation to ensure the correctness of outputs. ii) \textit{Agent adapter} may need additional memory or external data store to record the historical tool interfaces.

\end{itemize}

\vspace{0.5em}\noindent \textbf{Known uses:} 

\begin{itemize}
    %\item \textit{LangSmith}~\footnote{\url{https://www.langchain.com/langsmith}}. LangSmith provides a DevOps platform for foundation model applications.

    \item \textit{AutoGen}~\footnote{\url{https://microsoft.github.io/autogen/docs/tutorial/tool-use}}. Users can register different tools in the agent, specifying the usage description. Registered tools will be leveraged by the agent during a conversation with user.
    
    %and use them during a conversation with the agent.

    \item \textit{Apple Intelligence}~\footnote{\url{https://www.apple.com/apple-intelligence/}}. \textit{Apple Intelligence} can support writing, image generation, schedule management across different products and applications. For instance, it can capture the entities in users' photo library and create emoji.
    % photo library -> emoji

    \item \textit{Semantic Kernel}~\footnote{\url{https://learn.microsoft.com/en-us/semantic-kernel/agents/plugins/?tabs=Csharp}}. Semantic Kernel can orchestrate agents and plugins to extend agents' skills. Plugins need to provide semantic description (e.g. input, output, side effects) for agents to understand.

    \item Yang et al.~\cite{yang2024swe} devise SWE-agent that can provide agent-computer interfaces, enabling foundation model-based agents to process code commands and resolve software engineering tasks.

\end{itemize}

\vspace{0.5em}\noindent \textbf{Related patterns:}

\begin{itemize}
    \item \textit{Prompt/response optimiser.} \textit{Prompt/response optimiser} can improve users' inputs, and the optimised prompts can be sent to other agents for goal achievement, while \textit{agent adapter} focuses more on the utilisation of external tools.

    \item \textit{Tool/agent registry.} \textit{Tool/agent registry} records the available external tools, while \textit{agent adapter} can convert the interface of selected tools into agent-friendly format.
\end{itemize}

\subsection{Agent Evaluator}

\vspace{0.5em}\noindent \textbf{Summary:} Agent evaluator can perform testing to assess the agent regarding diverse requirements and metrics.

\vspace{0.5em}\noindent \textbf{Context:} Within an agent, the underlying foundation model and a series of components coordinate to conduct reasoning and generate the responses given users' prompts.

\vspace{0.5em}\noindent \textbf{Problem:} How to assess the performance of agents to ensure they behave as intended?

\vspace{0.5em}\noindent \textbf{Forces:} 
\begin{itemize}
    %\item \textit{Response accuracy.}

    \item \textit{Functional suitability guarantee.} Agent developers need to ensure that a deployed agent operates as intended, providing complete, correct, and appropriate services to users.
    
    \item \textit{Adaptability improvement.} Agent developers need to understand and analyse the usage of agents in specific scenarios, to perform suitable adaptations.

\end{itemize}

\begin{figure}[!ht]
    \centering
    \includegraphics[width=0.56\columnwidth]{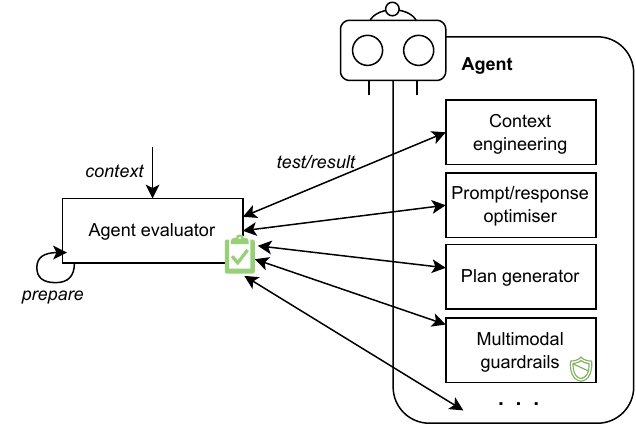}	
    \caption{Agent evaluator.}
    \label{fig:evaluator}
\end{figure}

\vspace{0.5em}\noindent \textbf{Solution:} Fig.~\ref{fig:evaluator} presents a simplified graphical representation of \textit{agent evaluator}. Developers can deploy evaluator to assess the agent regarding responses and reasoning process at both design-time and runtime. Specifically, developers need to build up the evaluation pipeline, for instance, by defining specific scenario-based requirements, metrics and expected outputs from agents. Given particular context, the \textit{agent evaluator} prepares context-specific test cases (either searching from external resources or generating by itself), and performs evaluation on the agent components respectively. The evaluation results provide valuable feedback such as boundary cases, near-misses, etc., while developers can further fine-tune the agent or employ corresponding risk mitigation solutions, and also upgrade the evaluator based on the results.

\vspace{0.5em}\noindent \textbf{Consequences:} 

Benefits:
\begin{itemize}
    \item \textit{Functional suitability.} Agent developers can learn the agent's behavior, and compare the actual responses with expected ones through the evaluation results.

    \item \textit{Adaptability.} Agent developers can analyse the evaluation results regarding scenario-based requirements, and decide whether the agent should adapt to new requirements or test cases.
    
    \item \textit{Flexibility.} Agent developers can define customised metrics and the expected outputs to test a specific aspect of the agent.
\end{itemize}

Drawbacks:

\begin{itemize}
    \item \textit{Metric quantification.} It is difficult to design quantified rubrics for the assessment of software quality attributes.

    \item \textit{Quality of evaluation.} The evaluation quality is dependent on the prepared test cases.
\end{itemize}

\vspace{0.5em}\noindent \textbf{Known uses:} 

\begin{itemize}
    \item \textit{Inspect}\footnote{\url{https://ukgovernmentbeis.github.io/inspect_ai/}}. UK AI Safety Institute devised an evaluation framework for large language models that offers a series of built-in components, including prompt engineering, tool usage, etc.

    \item \textit{DeepEval}\footnote{\url{https://docs.confident-ai.com/}}. DeepEval incorporates 14 evaluation metrics, and supports agent development frameworks such as LlamaIndex, Hugging Face, etc.

    \item \textit{Promptfoo}\footnote{\url{https://github.com/promptfoo/promptfoo}}. Promptfoo can provide efficient evaluation services with caching, concurrency, and live reloading, and also enable automate scoring based on user-defined metrics.

    \item \textit{Ragas}\footnote{\url{https://github.com/explodinggradients/ragas}}. Ragas facilitates evaluation on the RAG pipelines via test dataset generation and leveraging LLM-assisted evaluation metrics.
    
\end{itemize}

\vspace{0.5em}\noindent \textbf{Related patterns:} \textit{Agent evaluator} can be configured and deployed to assess the performance of other pattern-oriented agent components during both design-time and runtime.

\section{Lessons Learned From the Pattern Catalogue}
\label{sec:discussion}

In this section, we first propose a decision model for selecting the 18 identified patterns to enhance the development of foundation mode based agents. Afterwards, we would like to share our experiences on the application of certain patterns in research projects as case studies. Finally, we discuss several insights we learned during the pattern collection process and previous research works.

\begin{figure*}[t]
    \centering
    \includegraphics[width=\textwidth]{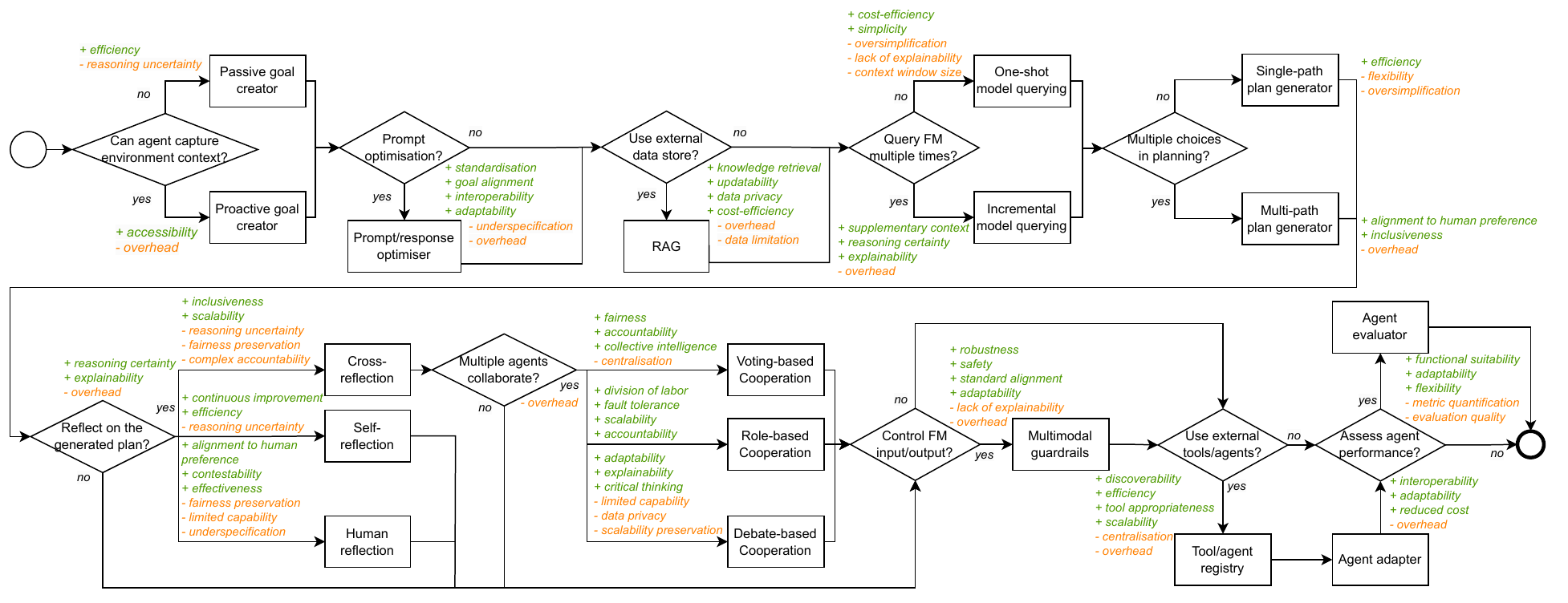}	
    \caption{Decision model for agent design pattern catalogue.}
    \label{fig:decision}
\end{figure*}

\subsection{Decision Model}
Fig.~\ref{fig:decision} illustrates a decision model which visualises the selection process of different patterns for practitioners. In particular, for a design problem, each decision can map to the corresponding solution space, which is regarded as the alternative pattern(s) for those in the opposite decision's solution space. Further, within the same solution space, there can be multiple patterns serving as complements of each other. The strengths and trade-offs of each pattern are highlighted in green and orange colour respectively. Please note that the decision model omits the common attributes for alternative patterns as either decision will incorporate these attributes as consequences, and it centralises the shared attributes for complementary patterns for brevity. In addition, there is no specific selection or application sequence, considering the patterns aim to facilitate the design and development of different architectural components, which can be implemented via a decoupling strategy. A brief explanation is provided as follows.

If the agent is expected to capture users' environmental information as supplementary context, \textit{proactive goal creator} can be applied to analyse users' goals based on the data captured by a series of sensors. In this case, the agent can also serve users with specified disabilities. Whilst, \textit{passive goal creator} can provide a simple and efficient dialogue interface to interact with users. \textit{Prompt/response optimiser} can enhance the goal alignment by refining users' instructions into standardised prompts. Meanwhile, the agent can retrieve more information from external knowledge base via \textit{retrieval augmented generation} whenever a component requires additional context. Besides, a component can query the incorporated foundation model for a single purpose multiple times (\textit{incremental model querying}) based on either user-specified requirements or system configuration, to provide supplementary context and hence improving model reasoning certainty, or just query once (\textit{one-shot model querying}) due to limited budget for model calling.

For a particular goal, agents can create a linear plan via \textit{single-path plan generator} for efficiency, or a complex plan in which each step has different options via \textit{multi-path plan generator} to ensure inclusiveness and alignment to human preferences. A generated plan can be assessed via multiple solutions (i.e. \textit{self-reflection, cross-reflection} and \textit{human reflection}) for ensuring plan correctness and feasibility, and improving agent reasoning certainty and explainability, while each solution has respective strengths and trade-offs. Further, multiple agents can be employed for reflection, and they can interact with each other in terms of \textit{voting-based, role-based} and \textit{debate-based} cooperation schemes.

When the underlying foundation model is queried, malicious inputs can affect the reasoning process, and the model may continuously learn and generate skewed outputs. \textit{Multimodal guardrails} can provide a layer between the foundation model and other components by inspecting the model inputs and outputs through both rule-based and AI-based examination. In case that external tools or agents are leveraged for certain tasks, a \textit{tool/agent registry} can enhance discoverability, and ensure the efficiency and appropriateness of tools. \textit{Agent adapter} guarantees the interoperability and adaptability of an agent to call external services whilst reducing the cost for manual development and maintenance. Finally, \textit{agent evaluator} can be utilised to assess the agent's functional suitability before release.

\subsection{Pattern Application}

In this section, we would like to discuss and share our experiences of applying the identified patterns in research projects. Researchers joined a project in which we are designing and developing an agent platform for creating and operating bespoke foundation model based agents with continuous learning capability. Specifically, the platform is divided into multiple products and components where different patterns can be applied.

Universal Task Assistant\footnote{\url{https://apputa.online/}}, developed by Xie et al., can help users discover and learn the usage of different mobile apps. This product applied \textit{proactive goal creator} to capture the mobile UI screen and detect the included elements, then it can perform actions to complete the user's tasks. Cheng et al. proposed an AI chain integrated development environment, Prompt Sapper~\cite{sapper}, for practitioners to properly and seamlessly develop their own FM-based AI chain services. In particular, the Prompt Sapper co-pilot implemented \textit{passive goal creator} for eliciting users' requirements, \textit{prompt/response optimiser} for refining users' task descriptions and generating AI chain skeleton, and \textit{tool/agent registry} for managing all available artifacts. Shamsujjoha et al.~\cite{shamsujjoha2024towards} further explored \textit{multimodal guardrails} by devising a taxonomy of guardrails, which provides comparative analysis of diverse design options in terms of the actions, targets, scopes, rules, autonomy, modalities, and underlying techniques when incorporating guardrails in foundation model based systems. Whilst, Xia et al.~\cite{xia2024towards} demonstrated an AI system evaluation framework, categorising the evaluation of foundation model based applications into system-level and component-level, each granularity has respective testing methods and benchmarking. The framework can offer guidance for the application of \textit{agent evaluator}.

\subsection{Discussion}

%This section discusses several insights based on our experiences during the pattern collection process and previous research projects.

\textbf{Integration with extant patterns.} Integrating different patterns can help structure and develop comprehensive and trustworthy agents. In particular, the proposed pattern catalogue can be applied together with responsible AI patterns~\cite{RAIbook} to ensure the agents behave in a responsible manner. For example, \textit{bill of materials registry} can record the procurement of components such as guardrails, prompt/response optimiser, etc., offering a complete supply chain of foundation model-based agent development. Whilst, \textit{black box} can be implemented to collect the inputs and outputs of foundation model-based agents at runtime. If any abnormal behaviour is detected or an audit is needed, the recorded information can provide evidence for the accountability process. Further, the collaboration patterns among agents can refer to existing social learning evolution models~\cite{9772408}. For instance, selecting suitable topologies and implementing effective control mechanisms can enhance the design of multi-agent workflows. Agents might also benefit from imitative learning in \textit{debate-based cooperation}, enabling dynamic adaptation and knowledge transfer. In addition, a set of voting mechanisms can be leveraged via blockchain smart contracts~\cite{10.1145/3551902.3564802} for \textit{voting-based cooperation}, to ensure transparent and secure interactions between agents. 

% topology and control mechanisms
% imitating and learn from each other

%Fed-ensemble: Ensemble Models in Federated Learning for Improved Generalization and Uncertainty Quantification
%Ensemble Federated Learning: An approach for collaborative pneumonia diagnosis

\textbf{Compliance with regulations and standards.} Preserving the alignment of agents with both international and domestic regulations and standards should be noted as a fundamental factor for developers to provide agent services in different countries and regions. European Parliament has approved the Artificial Intelligence Act\footnote{\url{https://artificialintelligenceact.eu/}} which focuses on four risk categories of AI applications. NIST released a draft publication for managing the risks of generative AI\footnote{\url{https://www.nist.gov/itl/ai-risk-management-framework}}, and ISO published an international standard for implementing and maintaining AI management systems~\cite{ISO42001}. Future work can extract and analyse the requirements within each regulation and standard, and map the proposed pattern catalogue to the requirements. In particular, we are adopting the concept of DevOps into FM-based agents, where both agent creation and operation information will be recorded for further auditing. Meanwhile, the agent workflow generation also requires inspection that whether it adheres to domain specifications.

%eu ai act, nist, iso standard
%mapping between patterns and regulations

\textbf{Evaluation of foundation model-based agents.} Evaluations on agents and the underlying foundation models are significant to ensure they behave as intended. %UK AI Safety Institute devised an evaluation framework for large language models\footnote{\url{https://ukgovernmentbeis.github.io/inspect_ai/}}. 
The majority of pattern benefits and drawbacks are software quality attributes, which still require quantification for fine-grained metrics and rubrics. For instance, accountability can be further divided into three criteria of responsibility, auditability, and redressability, and each criterion has its own process, resource, and product metrics~\cite{xia2023principles}. Proper quantification can promote the evaluation of agents and validate the effectiveness of applied patterns. Further, authors are also exploring real-time exception identification and handling in agent workflow execution, which includes rationalising task dependencies, external tool API version matching, etc.

%Please note that the evaluation of foundation model-based agents can be categorised into system-level and component-level~\cite{xia2024towards}, while each granularity has respective testing methods and benchmarking.

% RAI pattern catalogue

% blockchain patterns for voting

% compliance to laws

% quantification of benefits/drawbacks

\section{Conclusion}
\label{sec:conclusion}

Foundation model based agents are gaining increasing attention in various domains to intellectualise and automate the business process. However, practitioners are troubled by architectural challenges to design agents. Our previous work demonstrates a reference architecture to present an overview of agent design~\cite{lu2024responsible}, while in this study, we scrutinise the forces, solutions, and trade-offs of 18 patterns. The pattern catalogue is provided as a holistic guidance for architects to better design and develop foundation model-based agents. In our future work, we will study how to apply the pattern catalogue with existing patterns to preserve the trustworthiness of agents, and further explore the architecture decisions that are related to foundation model-based agents.


% \bibliographystyle{IEEEtran}
% \bibliography{bibliography}

\end{document}